\documentclass[final,5p,times,twocolumn,colorlinks=true,linkcolor=blue,citecolor=blue,urlcolor=blue,hyphens,spaces,nonatbib]{elsarticle}

\usepackage[backend=biber,style=apa]{biblatex}
\DeclareLanguageMapping{american}{american-apa}
\usepackage{amssymb}
\usepackage{amsmath}

\usepackage{orcidlink}

\usepackage{tabularx}

\usepackage{array}

\usepackage{booktabs} 
\usepackage{multirow} 

\usepackage{longtable}
\usepackage{adjustbox}

\usepackage{subcaption}

\usepackage{ragged2e}
\newcolumntype{L}[1]{>{\raggedright\arraybackslash}p{#1}}   
\newcolumntype{C}[1]{>{\centering\arraybackslash}m{#1}}     
\newcolumntype{R}[1]{>{\raggedleft\arraybackslash}p{#1}}    
\newcolumntype{J}[1]{>{\justifying\arraybackslash}p{#1}}    

\journal{arXiv}

\begin{document}

\begin{frontmatter}



\title{Enhancing Sentiment Classification and Irony Detection in Large Language Models through Advanced Prompt Engineering Techniques}


\author{Marvin Schmitt\texorpdfstring{\corref{Marvin}}{}\orcidlink{0009-0009-7780-4537}}
\author{Anne Schwerk\texorpdfstring{\corref{Anne}}{}\orcidlink{0000-0002-5014-5289}}
\author{Sebastian Lempert\texorpdfstring{\corref{Sebastian}}{}\orcidlink{0000-0002-5214-5944}}

\cortext[Sebastian]{Corresponding author, \textit{E-mail address}: \href{mailto:sebastian.lempert@iu.org}{sebastian.lempert@iu.org}}
\cortext[Marvin]{\textit{E-mail address}: \href{mailto:marvin.schmitt@iu.org}{marvin.schmitt@iu.org}}
\cortext[Anne]{\textit{E-mail address}: \href{mailto:anne.schwerk@iu.org}{anne.schwerk@iu.org}}

\affiliation{organization={IU International University of Applied Sciences},
            addressline={Juri-Gagarin-Ring 152}, 
            city={Erfurt},
            postcode={99084}, 
            state={Thuringia},
            country={Germany}}

\begin{abstract}
This study investigates the use of prompt engineering to enhance large language models (LLMs), specifically GPT-4o-mini and gemini-1.5-flash, in sentiment analysis tasks. It evaluates advanced prompting techniques like few-shot learning, chain-of-thought prompting, and self-consistency against a baseline. Key tasks include sentiment classification, aspect-based sentiment analysis, and detecting subtle nuances such as irony. The research details the theoretical background, datasets, and methods used, assessing performance of LLMs as measured by accuracy, recall, precision, and F1 score. Findings reveal that advanced prompting significantly improves sentiment analysis, with the few-shot approach excelling in GPT-4o-mini and chain-of-thought prompting boosting irony detection in gemini-1.5-flash by up to 46\%. Thus, while advanced prompting techniques overall improve performance, the fact that few-shot prompting works best for GPT-4o-mini and chain-of-thought excels in gemini-1.5-flash for irony detection suggests that prompting strategies must be tailored to both the model and the task. This highlights the importance of aligning prompt design with both the LLM’s architecture and the semantic complexity of the task.
\end{abstract}

\begin{keyword}
sentiment analysis \sep irony detection \sep large language models (LLMs) \sep prompt engineering



\end{keyword}

\end{frontmatter}



\section{Introduction}

Large language models (LLMs) like GPT-4 have demonstrated remarkable capabilities across natural language processing (NLP) tasks, including sentiment analysis \autocite{zhangSentimentAnalysisEra_NAACL_2024}. However, effectively harnessing these models often depends on how they are prompted to produce a good answer of high quality. Prompt engineering – the craft of designing input prompts to guide the model’s outputs – has emerged as a key factor in unlocking LLMs’ full potential \autocite{schulhoffPromptReportSystematic_arXiv_2024,chenUnleashingPotentialPrompt_arXiv_2023,marvinPromptEngineeringLarge_ICDICI_2023}. Recent advances propose sophisticated prompting techniques (e.g. providing few-shot examples or instructing the model to reason step-by-step) to improve performance beyond naive zero-shot usage.

This study focuses on analyzing how such advanced prompt engineering strategies impact LLM performance on sentiment analysis tasks, specifically evaluating two cutting-edge models (OpenAI’s GPT-4o-mini and Google’s gemini-1.5-flash) across a spectrum of sentiment analysis challenges. We compare a baseline prompt (simple zero-shot instruction) against advanced prompting approaches – including few-shot prompting, chain-of-thought (CoT) reasoning, and self-consistency – and quantitatively assess their impact using standard classification metrics like accuracy and F1-score. By systematically varying the prompting strategy, we aim to determine to what extent prompt engineering can optimize sentiment classification outcomes.

To thoroughly evaluate these effects, we consider multiple representative datasets covering both straightforward and nuanced sentiment analysis scenarios. The experimental tasks span:

\begin{itemize}
    \item Binary sentiment classification: Stanford Sentiment Treebank (SST-2) movie reviews (English), labeled positive vs. negative
    \item Multilingual three-class sentiment: SB-10k corpus of 10,000 German tweets labeled positive, neutral, or negative
    \item Aspect-based sentiment analysis (ABSA): SemEval-2014 ABSA challenge dataset (customer reviews in domains like laptops and restaurants) annotated with sentiment for specific aspects in each text. 
    \item Irony detection: SemEval-2018 Task 3 dataset of 3,000 English tweets, each manually annotated as ironic or not ironic.
\end{itemize}

By evaluating prompt techniques on this diverse set of sentiment analysis problems – from basic polarity classification to aspect-level opinion mining and recognizing irony – we ensure a comprehensive analysis of prompt engineering’s impact. The inclusion of both English and non-English content (e.g. German tweets in SB-10k) further tests each model’s adaptability to multilingual prompts. Through this broad experimental scope, our work investigates whether carefully engineered prompts can consistently boost LLM accuracy across varying sentiment tasks, and how the effectiveness of these techniques may differ by task type and by model.

\subsection{Motivation}

Sentiment analysis is a long-standing challenge in NLP, and certain sub-tasks remain difficult for traditional approaches, as for example n-grams \autocite{jimRecentAdvancementsChallenges_NLPJ_2024}. For instance, detecting aspect-based sentiments within text (identifying sentiment toward specific attributes in a review) and recognizing subtle linguistic nuances like irony or sarcasm are complex problems that often require contextual reasoning beyond surface-level cues \autocite{jimRecentAdvancementsChallenges_NLPJ_2024}. Contemporary LLMs bring new capabilities to these problems – they possess broad world knowledge and the ability to perform reasoning in context – yet leveraging these capabilities is not trivial.

While fine-tuned models have historically dominated sentiment analysis, the rise of powerful generative LLMs suggests we might obtain strong performance without task-specific training if we can prompt the model effectively \autocite{zhouComprehensiveEvaluationLarge_arXiv_2024}. Prompt engineering offers a flexible, data-efficient way to adapt a single LLM to multiple sentiment tasks by simply varying its input instructions, potentially obviating the need for collecting large labeled datasets or training separate models for each sentiment domain.

Despite the explosion of LLM applications, there is a notable research gap on how advanced prompting techniques specifically affect sentiment analysis performance \autocite{sunHarnessingDomainInsights_KBS_2024}. Prior works have extensively benchmarked different LLMs on general NLP tasks, but few have isolated the contribution of prompt design on sentiment tasks. This gap is significant: sentiment analysis often involves subjective and context-dependent judgments that might benefit from prompts which encourage the model to think (e.g. via CoT) or see more examples (few-shot). We are motivated to address whether and how specialized prompts can reliably enhance LLM accuracy in sentiment analysis, as opposed to using a generic prompt. Understanding this would both advance the theory of prompt-based learning and inform practical deployment of LLMs for sentiment tasks. From an applied perspective, even modest gains in sentiment classification accuracy can be impactful. For example, in customer service applications, accurately detecting a customer’s sentiment and any sarcastic tone in their message is crucial for generating appropriate, empathetic responses. Prompt strategies that significantly improve irony detection or aspect understanding could be directly leveraged to build more sensitive sentiment analysis systems in marketing, finance, or social media monitoring.

In summary, the motivation of this work is to explore prompt engineering as a means to bridge the gap between general-purpose LLMs and the specialized demands of sentiment analysis. By experimentally quantifying improvements due to prompts, we seek to contribute insights on maximizing LLM performance in affective computing contexts without additional model fine-tuning. This has important theoretical implications for how language models interpret prompts, and practical implications for rapidly adapting AI to real-world sentiment understanding problems.

\subsection{Research question} \label{rq}

Building on the above motivations, our study is guided by the following research question:

\textit{RQ: To what extent can advanced prompt engineering techniques (such as few-shot examples, CoT prompting, and self-consistency) improve the accuracy and F1-score of LLMs on sentiment analysis tasks compared to a baseline zero-shot prompt?}

We ask this in the context of both simple sentiment classification and more nuanced sentiment detection (e.g., irony).

While this study primarily focuses on performance gains, it also considers additional aspects of prompt engineering. Specifically, we investigate variations in these effects across different LLM architectures (e.g., OpenAI GPT-4o-mini vs. Google gemini-flash-1.5). Furthermore, we explore the efficacy of specific prompting strategies for distinct sentiment analysis tasks (e.g., reasoning-based prompts (CoT) for sarcasm detection) and assess whether techniques such as self-consistency enhance the stability and robustness of model predictions. 


\section{Related work}

\subsection{LLMs in sentiment analysis}

The advent of generative LLMs has prompted researchers to evaluate their capabilities on sentiment tasks traditionally handled by smaller, fine-tuned models.

\textcite{zhouComprehensiveEvaluationLarge_arXiv_2024} conducted a comprehensive evaluation of multiple LLMs on aspect-based sentiment analysis (ABSA) across 13 datasets. They found that with appropriate prompting (in-context demonstrations), large models can achieve state-of-the-art performance on ABSA, in some cases rivaling or surpassing fine-tuned smaller models without any task-specific training. This underscores the promise of prompt-based learning for sentiment tasks.

In contrast, \textcite{filipFinetuningMultilingualLanguage_arXiv_2024} indicate with regard to ABSA that lightly fine-tuned medium-sized models can outperform in-context prompting on fine-grained sentiment tasks.

In the financial domain, \textcite{luoPretrainedLargeLanguage_arXiv_2024} showed that adapting a pre-trained LLM through light supervised fine-tuning drastically improved financial news sentiment classification, outperforming prior state-of-the-art algorithms. Notably, even a relatively small 7B-parameter LLaMA 2 model, when properly adapted, exceeded the accuracy of specialized models on detecting positive/negative sentiment in stock news.

These studies illustrate that LLMs possess strong capabilities for sentiment analysis which can be unlocked either by careful prompting or minimal fine-tuning.

Several works have specifically explored targeted or fine-grained sentiment tasks with LLMs, highlighting both their strengths and the importance of prompt design.

\textcite{jurosLLMsTargetedSentiment_WASSA_2024} investigated targeted sentiment analysis of news headlines (sentiment toward a particular entity in the headline) using GPT models. They reported that LLMs outperformed traditional fine-tuned encoder-based models on datasets with more descriptive annotations, thanks to LLMs’ broad knowledge and reasoning. Moreover, they found that the level of prompt prescriptiveness markedly affected performance: providing more elaborate instructions or few-shot examples improved F1 scores and calibration of the model’s confidence, though there was an optimal level beyond which returns diminished. This result points to the delicate balance in prompt engineering – overly simple prompts underutilize the model, while overly complex prompts can introduce confusion or noise.

\textcite{stigallLargeLanguageModels_ACMSE_2024} provide a related perspective, comparing the performance of various LLMs on emotion and sentiment classification tasks. Their work introduced a fine-tuned small model “EmoBERT\textsubscript{Tiny}” specialized for emotion detection, which actually outperformed some larger pre-trained LLMs on sentiment/emotion benchmarks. This finding suggests that without proper prompting or adaptation, a generic LLM may still lag behind a dedicated model, reinforcing the need for effective prompt engineering (or fine-tuning) to get the best results from LLMs.

\subsection{Prompt engineering and affective computing}

With LLMs becoming prevalent, researchers have started examining how prompt formulation influences their performance in affective computing tasks (sentiment, emotion, toxicity, sarcasm).

\textcite{aminPromptSensitivityChatGPT_arXiv_2024} present a focused study on the prompt sensitivity of ChatGPT (GPT-3.5 and GPT-4) for sentiment analysis, toxicity detection, and sarcasm detection. By systematically varying prompt wording, structure, and generation parameters, they demonstrated that the model’s performance can vary significantly under different prompts, even when the core task remains the same. Certain phrasings or instructions yielded substantially better accuracy, whereas others led to declines, highlighting that prompt design is an essential lever for optimizing LLM behavior. They also explored how decoding settings (like temperature in generation) affect outcomes, noting a trade-off between creativity and consistency that practitioners must manage.

\textcite{yaoSarcasmDetectionStepbyStep_arXiv_2024} propose “SarcasmCue”, a framework with four prompting methods (chain of contradiction, graph of cues, bagging of cues, tensor of cues) to help LLMs detect sarcasm by considering both sequential and non-sequential cues. Their experiments show: (1) advanced models like GPT-4 and Claude 3.5 perform best with CoC and GoC prompts; (2) the ToC method greatly boosts smaller LLM performance; and (3) their framework consistently beats the previous state-of-the-art across four datasets, proving its effectiveness and robustness.

\textcite{zhangSarcasmBenchEvaluatingLarge_arXiv_2024} introduce “SarcasmBench”, a benchmark for evaluating LLMs on sarcasm detection, testing eleven leading LLMs and eight pre-trained language models across six datasets using various prompting methods. Their key findings are: (1) current LLMs perform worse than supervised PLMs for sarcasm detection, indicating more work is needed; (2) GPT-4 outperforms other LLMs by a significant margin; and (3) few-shot input/output prompting outperforms zero-shot input/output prompting and few-shot CoT.

The importance of prompt design is further echoed by surveys of LLM applications in sentiment analysis. 

\textcite{yangLargeLanguageModelsMeetTextCentric_arXiv_2024} provide a survey on text-centric multimodal sentiment analysis in the LLM era, noting that while models like ChatGPT open new possibilities, it remains unclear how to best adapt them to complex multimodal sentiment tasks, and that careful prompt and instruction tuning is likely required.

\textcite{krugmannSentimentAnalysisAge_CNS_2024} examined sentiment analysis in the age of generative AI and found that even in zero-shot settings, LLMs like GPT-4 can match or exceed the accuracy of domain-specific sentiment classifiers. They emphasize, however, that the characteristics of the input text and the prompt heavily influence performance: for example, longer and more content-rich texts allowed the LLMs to shine, whereas very short or informally written inputs saw the LLMs struggle somewhat. Notably, \textcite{krugmannSentimentAnalysisAge_CNS_2024} also reported that LLMs (especially an instruction-tuned model like LLaMA 2) can provide human-like explanations for their sentiment decisions, which is a valuable byproduct of techniques like CoT prompting that elicit the model’s reasoning.

\subsection{Summary of related work and positioning of this work}

In summary, existing work establishes that LLMs are powerful tools for sentiment analysis across domains (from customer reviews to news and finance) when used appropriately. Proper prompt engineering appears repeatedly as a decisive factor: it can determine whether an LLM merely “competes” with older models or decisively surpasses them \autocite{krugmannSentimentAnalysisAge_CNS_2024}.

Our research builds on this foundation by providing a focused, experimental look at advanced prompting techniques (few-shot, CoT, self-consistency) across multiple sentiment tasks. This bridges the gap between broad evaluations of LLMs on sentiment tasks and a targeted analysis of how to prompt them best for optimal performance.

\section{Background} \label{sec_background}

\subsection{Prompting}

Before providing the methodological details, we provide a brief background on the relevant areas of prompting, prompt engineering, and sentiment analysis.

First, a prompt serves as the interface between users and LLMs. It is an instruction, possibly accompanied by specific contextual information, that is given to a model to elicit a particular response. Prompts can vary widely, ranging from simple, everyday expressions to complex, nested instructions \autocite{liuPretrainPromptPredict_arXiv_2021}.

\textcite{marvinPromptEngineeringLarge_ICDICI_2023} further elucidate these fundamental principles by breaking down a prompt into components such as instruction, input data, context, and an output directive. Specifying the desired output format, such as CSV or Markdown, can help structure the LLM-generated output. Style instructions are an additional variant of formatting directives, adjusting the output's style without altering its structure. Finally, supplementary information can also be part of a prompt to contextualize the output. When instructing to draft an email, details like name and position might be added as system instructions to create an appropriate signature \autocite{schulhoffPromptReportSystematic_arXiv_2024}.

\subsection{Prompt engineering}
Second, the concept of prompt engineering pertains to the design and conceptualization of prompts. Adhering to various rules and nuances in prompt formulation or handling can significantly enhance output performance and quality \autocite{whitePromptPatternCatalog_arXiv_2023}. Additionally, LLMs can be "programmed" using prompt engineering. For example, a prompt can be structured to prompt the model to ask the user multiple questions until sufficient information is gathered to generate a specific output, such as a piece of code \autocite{whitePromptPatternCatalog_arXiv_2023}.

In addition to the advanced prompt engineering techniques examined in this research, it is essential to acknowledge certain foundational principles that, while not classified as prompt engineering approaches within the scope of this study, still play a critical role in effective prompt design. These include the importance of clear and precise prompt formulation, ensuring that the model receives unambiguous instructions tailored to the specific task \autocite{chenUnleashingPotentialPrompt_arXiv_2023}. Furthermore, role prompting, in which the model is explicitly assigned a role (e.g., "you are a sentiment analysis expert"), can help guide the model’s responses by providing contextual framing \autocite{chenUnleashingPotentialPrompt_arXiv_2023}. While these principles are not considered advanced techniques in this work, they serve as fundamental best practices in prompt creation and contribute to establishing a strong baseline for subsequent optimizations.

\subsection{Sentiment analysis}

Third, since this study centers on the field of sentiment analysis, it is essential to highlight its broad range of applications. sentiment analysis is employed across various domains, including the analysis of customer feedback, market monitoring, the optimization of customer service, and the evaluation of public opinion in political and social contexts. Owing to its versatility, sentiment analysis serves as a key enabler of data-driven decision-making and facilitates more personalized and context-sensitive responses to individual needs \autocite{swetaSentimentAnalysisIts_Springer_2024}.

\textcite{swetaSentimentAnalysisIts_Springer_2024} highlights different areas within sentiment analysis, including binary classification (positive/negative), multi-class classification (including neutral), and aspect-based analyses that focus on specific features or attributes of an object. Further specialized areas encompass irony detection, temporal analysis of opinion shifts, domain-specific adaptations, and fine-grained analyses that capture the strength or intensity of opinions.

In this study, we focus on selected areas of sentiment analysis, as they represent various dimensions of text analysis and enable a nuanced assessment of the performance of LLMs.
Specifically, the inclusion of classical sentiment classification (SC), ABSA, and multi-faceted analysis of subjective texts (MAST) permits an in-depth exploration of both the versatility and limitations of LLMs and individual prompting approaches in the context of sentiment analysis. While SC is concerned with categorizing texts into predefined sentiment categories, ABSA allows a more detailed examination of specific aspects mentioned within the text. MAST, on the other hand, encompasses more complex tasks such as irony detection, which require a deeper understanding of linguistic nuances and contextual interpretation \autocite{zhangSentimentAnalysisEra_NAACL_2024, wangInvestigatingImpactPrompt_JMIR_Formative_Research_2024}.

\section{Experimental design}

\subsection{Prompt engineering approaches}

The selection of shot prompting, CoT, and self-consistency in this study is grounded in their demonstrated effectiveness for complex, context-dependent tasks such as sentiment analysis. As \textcite{sahooSystematicSurveyPrompt_arXiv_2024} emphasize, prompt engineering strategies must be adapted to the specific domain, with sentiment analysis particularly requiring reasoning over subtle and nuanced language. These approaches have consistently shown substantial performance gains in the literature \autocite{zhangSentimentAnalysisEra_NAACL_2024, weiChainofthoughtPromptingElicits_NIPS_2022, luoSelfAttentionTransformersDriving_ICEICT_2023}, offering structured reasoning processes that help reduce inconsistencies and infer implicit meanings. Their domain alignment and empirical success make them a well-justified focus for our investigation.

 Table~\ref{prompt_approaches} provides an overview of the prompting strategies employed in this study, summarizing the respective advantages and limitations of each approach. The complete set of prompts used throughout the study is presented in ~\ref{appendix:all_prompts}. 

\begin{table*}[!ht]
    \centering
    \begin{tabularx}{\linewidth}{|m{3cm}|X|X|}
        \hline
        \multicolumn{1}{|c|}{\textbf{Prompt approach}} & \multicolumn{1}{|c|}{\textbf{Advantages}} & \multicolumn{1}{|c|}{\textbf{Limitations}} \\
        \hline
        zero-shot & Efficient, no examples needed & Limited accuracy for complex tasks; struggles with subtle contexts \\ \hline
        One-/few-shot & Improves performance with examples; adapts to tasks & Risk of overfitting; depends on example quality; potential bias \\ \hline
        CoT & Enhances structured thinking; improves accuracy in complex tasks & Inefficient for simple tasks; requires large LLMs, increasing cost \\ \hline
        zero-shot-CoT & Combines zero-shot with CoT; reduces example dependency & Needs large LLMs; sensitive to question phrasing; possible inefficiencies \\ \hline
        self-consistency & Increases response reliability; enhances multi-path tasks & Needs multiple runs (higher resource use); risk of inconsistent outputs \\ \hline
    \end{tabularx}
    \caption{Overview of prompting approaches}
    \label{prompt_approaches}
\end{table*}

\subsection{Hypotheses and experimental setup} \label{Experimental_design}

As outlined in the central research question (Chapter \ref{rq}), this study examines the extent to which advanced PE techniques can enhance the performance of LLMs on SA tasks. The analysis spans both standard sentiment classification and more nuanced tasks such as irony detection. Accordingly, the experimental design is structured around three hypotheses, each addressing a specific aspect of the research question:
 
\begin{itemize} \label{hypotheses}
\item H1: The application of various prompt engineering techniques increases the accuracy and F1-score of LLMs in sentiment classification compared to baseline prompts.
\item H2: Precisely designed prompt engineering techniques enhance the ability of LLMs to detect aspect-based sentiments within a text compared to generic prompts, as measured by metrics such as accuracy and weighted F1-score.
\item H3: Well-structured prompts utilizing advanced techniques significantly improve accuracy and F1-score in the detection of ironic tweets compared to simpler prompts.
\end{itemize}

The experimental design follows the approach of \textcite{zhangSentimentAnalysisEra_NAACL_2024}, as their methodology provides a comprehensive and nuanced evaluation of the performance of LLMs across various sentiment analysis tasks. Their study incorporates both straightforward classification tasks and more complex analyses, such as ABSA and irony detection.

Building on this foundation, the experimental procedure adopted in this study follows a structured, multi-step process designed to systematically investigate different prompt-engineering techniques and their impact on LLM performance in the three specified areas of sentiment analysis. The impact of these techniques is assessed by comparing their performance against a baseline prompt, with accuracy and F1-score serving as key evaluation metrics. 

For the classification process (the LLM invocation), each text instance is submitted to the selected LLMs (GPT-4o-mini and gemini-1.5-flash) via API requests. To balance computational efficiency and statistical validity, a subset of 1,000 randomly sampled entries per dataset is used. To assess whether performance differences between prompting strategies (e.g., baseline vs. one-shot) are statistically significant, a bootstrap resampling procedure was applied. In this process, 1,000 bootstrap samples were drawn with replacement from the evaluation set, and the weighted F1-score was computed for each model within each sample. The resulting distribution of F1-score differences was then used to construct a 95 \% confidence interval. If this interval does not include zero, the observed performance difference is considered statistically significant. \ref{appendix:statTest} provides a detailed account of the results obtained through the bootstrap resampling procedure.

The experiment also considers language consistency: prompts are formulated in the same language as the dataset to prevent misinterpretations caused by language mismatches. The structured design progresses through the three analysis tasks — sentiment classification, ABSA, and irony detection — evaluating results for each prompt type across both models.

\subsubsection{Models} \label{chapter: Models}

For this experiment, the current model variants as of January 2025 (GPT-4o-mini from OpenAI and gemini-1.5-flash from Google) were deliberately selected, as they optimally meet the requirements for both performance and cost efficiency. The choice of these specific models is based on a strategic consideration: both variants offer powerful language understanding while being cost-optimized, enabling a resource-efficient execution of the experiment. This is particularly relevant in the context of a scientific study, where a balance between precise results and feasible cost structures is essential.

GPT-4o-mini is a more compact version of GPT-4, specifically designed for use cases requiring high performance without incurring the costs associated with larger variants \autocite{openaiGPT4oSystemCard2024,openaiGPT4oMiniAdvancing2024,openaiGPT4oMini2025}. It thus provides a solid foundation for testing the effectiveness of prompt engineering techniques in sentiment analysis.

At the same time, gemini-1.5-flash, the most balanced model in the current Gemini series (as of January 2025), allows for a comprehensive evaluation under an alternative architectural approach \autocite{googleGeminiModelsGemini2025}.

Both models provide dedicated Python libraries that seamlessly integrate into the Jupyter environment \autocite{openaiOpenaiOfficialPython2025,googleGooglegenaiGenAIPython2025}. These libraries facilitate interaction with the models by enabling a structured and efficient use of their respective APIs. Once installed and configured, queries can be executed directly from the code, ensuring a flexible and reproducible interaction with the language models \autocite{caelenDevelopingAppsGPT42024}.

From a technical perspective, both models utilize structured prompts, distinguishing between system prompts (which define the model’s role and behavior) and user prompts (which contain the specific classification request). The experimental setup follows this schema, with system instructions tailored to each subtask. To ensure consistency, the temperature parameter is reduced from the default value of 1 to 0.2, minimizing variability in model responses.

\subsubsection{Datasets}

For each area of the sentiment analysis examined in this study, a distinct dataset is utilized. In the case of sentiment classification, two different datasets are employed, covering both binary classification and multi-class sentiment classification. Additionally, a German-language dataset is included to evaluate the models' linguistic versatility. For the remaining two areas, ABSA and irony detection, one dataset per task is applied. An overview of the datasets is provided in Table \ref{table_datasets}.

\begin{table*}[!ht]
    \centering
    \begin{tabularx}{\linewidth}{|>{\raggedright\arraybackslash}m{2cm}|X|>{\centering\arraybackslash}m{2cm}|>{\centering\arraybackslash}m{2cm}|}
        \hline
        \multicolumn{1}{|c|}{\textbf{Name}} & \multicolumn{1}{c|}{\textbf{Description}} & \multicolumn{1}{c|}{\textbf{Usage}} & \multicolumn{1}{c|}{\textbf{Sources}} \\
        \hline
        SB10k & The corpus SB10k contains 9783 German tweets. Each tweet has sentiment annotations on tweet level by 3 human annotators, using sentiment classes positive, negative, neutral, mixed, and unknown. For the experiment, all tweets classified as mixed or unknown were removed. & Sentiment classification & \textcite{cieliebakTwitterCorpusBenchmark_SocialNLP_2017,geislingerAlienmasterSB10kDataset_HuggingFace_2024,paperswithcodeSB10kDataset_PapersWithCode_2024} \\
        \hline
        SST-2 /\linebreak SST binary & The Stanford Sentiment Treebank (SST) is a corpus with fully labeled parse trees that were parsed from 11,855 single sentences that were extracted from movie reviews. Each phrase has fine-grained sentiment annotations by 3 human annotators, using sentiment classes very negative, negative, neutral, positive, and very positive. The subset containing binary classifications using sentiment classes positive and negative (neutral sentences were discarded) is referred to as SST-2 or SST binary. & Sentiment classification & \textcite{socherRecursiveDeepModels_EMNLP_2013,jiangSST2SentimentAnalysisDataset_GitHub_2020,paperswithcodeSST2Dataset_PapersWithCode_2024} \\
        \hline
        SemEval 2014-ABSA Challenge Dataset & This dataset was introduced as part of the SemEval-2014 Task 4, which focused on ABSA. The dataset consists of customer reviews from domains such as restaurants and laptops. Each review is annotated to identify both the relevant aspects (e.g. “price”, “food” or “battery life”) and the associated sentiments (positive, negative or neutral). & ABSA & \textcite{pontikiAspectBasedSentimentAnalysis_SemEval_2014} \\
        \hline
        SemEval-2018-Irony Dataset & This dataset is designed to advance research in distinguishing ironic expressions, which can significantly affect sentiment analysis and natural language understanding. This dataset includes 3,000 English-language tweets that were labeled manually by three independent annotators. The corresponding labels of the tweets are binary categorized and indicate whether a tweet contains irony or not. & Irony detection & \textcite{vanheeIronyDetectionInEnglishTweets_SemEval_2018} \\
        \hline
    \end{tabularx}
    \caption{Datasets}
    \label{table_datasets}
\end{table*}

\subsubsection{Evaluation}

To establish a reference point, the experiment begins with a baseline prompt using a simple sentiment classification prompt (e.g., “Classify the following statement as positive, neutral, or negative”). This baseline serves as a benchmark for subsequent advanced prompting techniques, such as few-shot learning, CoT, and self-consistency. These advanced approaches are designed to improve the model’s ability to handle complex tasks like irony detection and ABSA.

Performance is assessed using classification metrics, including accuracy, precision, recall, and F1-score, with particular attention to class-specific performance (e.g., positive vs. negative sentiment). Following result documentation, a detailed analysis is conducted to identify systematic misclassifications and assess potential weaknesses in different prompting techniques. This provides deeper insights into model limitations and opportunities for further optimization. This rigorous documentation minimizes methodological biases and supports the validation and extension of findings.

Building upon the baseline prompt, more advanced prompts were designed and iteratively refined using the prompt engineering techniques outlined in Chapter \ref{sec_background}. The effectiveness of each prompting strategy is then systematically evaluated using standard classification metrics, including accuracy, precision, recall, and F1-score. These metrics not only provide an overall performance measure but also allow for direct comparisons between the different prompt types. Furthermore, to enable a more granular analysis, the metrics are computed separately for individual sentiment classes (e.g., positive, negative), thereby offering deeper insights into the model’s ability to distinguish between varying sentiment expressions.

\section{Results}

\begin{table*}[!ht]
\resizebox{\textwidth}{!}{%
    \begin{tabular}{|c|cc|c|c|}
    \hline
    \multirow{2}{*}{\textbf{Prompt approach}}
    & \multicolumn{2}{c|}{\textbf{Sentiment classification (H1)}}
    & \textbf{ABSA (H2)}
    & \textbf{Irony detection (H3)} \\
    \cline{2-5} 
    & \multicolumn{1}{c|}{\textbf{SST2}}
    & \textbf{SB10k}
    & \textbf{SemEval-2014}
    & \textbf{SemEval-2018} \\
    \hline
    GPT
    & \multicolumn{1}{c|}{few-shot 0.93 (+2 \%)}
    & few-shot 0.72 (+14 \%)
    & few-shot 0.85 (+2,4 \%)
    & few-shot 0.76 (+4 \%) \\
    \hline
    Gemini
    & \multicolumn{1}{c|}{CoT 0.95 (+12 \%)}
    & few-shot 0.61 (+15 \%)
    & CoT / self-consistency 0.83 (+2,5 \%)
    & CoT 0.6 (+46 \%) \\
    \hline
    \end{tabular}%
}
\caption{Overview of the best results (weighted F1) compared to the baseline approach}
\label{result_overview}
\end{table*}

Based on this factual presentation, the following analysis aims to uncover the specific limitations of each approach and to identify recurring patterns in misclassifications. The analysis investigates whether certain sentiment categories or linguistic nuances systematically pose greater classification challenges, thereby identifying key areas for potential optimization in future work. An overview of the key results is presented in Table \ref{result_overview}. A comprehensive and objective summary of all results is provided in \ref{appendix:all_results}. Furthermore, to support interpretation and enhance understanding, selected confusion matrices are provided in \ref{appendix:conf_m} and are referenced in the subsequent analysis.

\subsection{Shot prompting}

In the binary classification setting, the selection of the example in the one-shot approach appears to have only a moderate impact on overall performance, suggesting that the specific choice of the exemplar does not significantly affect classification outcomes.  In the context of binary classification, one might expect that performance improvements would primarily occur within the class represented by the selected example. However, an examination of the confusion matrix for irony detection under the gemini-1.5-flash model (see Figure~\ref{fig:cm_irony1}) reveals that, despite the example being drawn from the ironic class, the model also demonstrates improved recognition of the non-ironic (negative) class.

Furthermore, in multi-class classification tasks, such as those in the SB10k dataset or ABSA, the selection of the example has a more pronounced effect on model performance. During the baseline evaluation, it was already observed that both models struggled to classify neutral sentiments. LLMs tend to assign a certain polarity to generated responses, making neutral sentiment classification particularly challenging.

To counteract this tendency, the one-shot prompt for these datasets included an example from the neutral class. A closer examination of the results reveals that the inclusion of this single neutral exemplar encouraged the models to classify instances as neutral more frequently, thereby mitigating the initial bias toward polarized sentiment predictions. Using this approach under gemini-1.5-flash, the recall for the neutral class in the SB10k dataset increased from \textit{0.37 to 0.51}, demonstrating a statistically significant improvement in the model’s ability to accurately identify neutral sentiments (see Figure~\ref{fig:cm_sb10k}). 

In the few-shot approach employed in this experiment, two examples per class were provided for each specific subtask. However, additional examples were included for the neutral class to encourage the selection of neutral sentiments.

As shown in Table~\ref{result_overview}, this strategy led to the best performance across most subtasks compared to all other prompting approaches. In particular, for the SB10k dataset, GPT-4o-mini exhibited significant improvements of approximately 10 percentage points in both accuracy and weighted F1-score compared to the baseline approach. Notably, this improved performance was not solely driven by better classification of the neutral class, as observed in the one-shot approach. Instead, the examples had a clear impact on all three sentiment classes, resulting in enhanced F1-scores across the board.

However, the positive effect of multiple examples appears to be more pronounced for GPT than for gemini-1.5-flash. While both models benefited from few-shot prompting compared to one-shot prompting, GPT-4o-mini exhibited a substantially stronger performance gain. For example, in the SB10k dataset, GPTs F1-score increased from \textit{0.68 to 0.72}, whereas gemini-1.5-flash showed only a marginal improvement of \textit{0.01} percentage points, effectively making its performance indistinguishable from the one-shot approach. Regarding statistical significance, the majority of performance improvements observed under the few-shot approach are statistically significant. Notable exceptions include the ABSA task when using gemini-1.5-flash and the irony detection task when using GPT-4o-mini, where the observed improvements did not reach statistical significance.

\subsection{Chain-of-thought}

Based on the current state of research, the CoT approach was initially anticipated to achieve the highest performance. However, under GPT-4o-mini, the results indicate that both accuracy and F1-score slightly decline starting from the CoT approach compared to the baseline, with the observed differences failing to reach statistical significance. 

An advantage of the CoT approach in this context is that it allows greater transparency in the decision-making process of the model, as its reasoning steps can be explicitly traced.

The following example highlights a misclassification, thereby revealing a possible shortcoming in the CoT-based reasoning process (GPT-4o-mini, irony detection):

\begin{itemize}
    \item Tweet: “Jusuf Nurkic \#Bulls RT @JonathanTjarks: He clearly has a lot of talent. Kind of amazing that Denver got him and Gary Harris for McDermott.” (Gold label: no irony)
    \item CoT Explanation: “[Step(explanation=“The text discusses a trade in which Denver acquired Jusuf Nurkic and Gary Harris in exchange for McDermott. The phrase ‘kind of amazing’ could imply that the speaker finds the trade to be surprisingly beneficial for Denver, suggesting that they got a lot of talent in return for McDermott. This could be seen as ironic if the speaker believes that Denver got a much better deal than expected, highlighting a disparity between the perceived value of the players involved in the trade.”, output=‘Yes’)]”
\end{itemize}

This example illustrates that while the CoT approach produces a plausible line of reasoning that justifies the presence of irony, it ultimately leads to an incorrect conclusion. In contrast, the baseline approach correctly classified the text as “no irony”.

Under gemini-1.5-flash, however, it can be concluded that the CoT approach demonstrates the best results in comparison. In the SST2 dataset, this approach achieves a very high accuracy and F1-score of \textit{0.95}, meaning that nearly all sentiments in the dataset are correctly classified. In addition to this dataset, clear performance improvements are also observed in irony detection under gemini-1.5-flash, with statistically significant gains. This is particularly noteworthy in this subdomain, as gemini-1.5-flash has generally exhibited poor performance in irony detection prior to this approach.

Upon closer examination of the recall within individual classes or by analyzing the confusion matrix (see Figure~\ref{fig:cm_irony_CoT_gemini}), it becomes apparent that the performance for the negative class, i.e., “no irony”, plays a critical role.

While in the zero-shot approach (baseline), almost all texts are classified as ironic, the CoT approach results in a stronger classification of “no irony”. This is reflected in a low recall for the negative class of \textit{0.06} in baseline, while the recall can be improved to \textit{0.38} in the CoT approach.

The zero-shot-CoT approach frequently produces results in sentiment classification that fall below those of the baseline approach when evaluated using weighted F1 or Accuracy.  

\subsection{Self-consistency}

In contrast to previous discussed results, the self-consistency approach forms an exception. For GPT-4o-mini and the SST2 dataset, it is the only approach that exhibits decreased performance compared to the baseline approach. Figure~\ref{fig:cm_SST2_GPT} comparatively juxtaposes the classification matrix from the best approach (few-shot) with the worst approach (self-consistency) in GPT-4o-mini.

In this context, it is noteworthy that the proportion of false-negative classifications in the model is particularly pronounced. This results in low precision in the negative class. It is also interesting to observe that the model appears “confident” in these misclassifications. The self-consistency approach, therefore, frequently provides an incorrect outcome for the same classification task, despite the majority decision (statistical mode).

An interesting aspect of the self-consistency approach is the interaction between the mentioned “n” parameter, which refers to the iteration frequency for a query, and the general "temperature" parameter of LLMs (see Chapter \ref{chapter: Models}). In the course of the experiment, a lower parameter setting of \textit{0.2} was chosen to naturally produce more consistent model results. Additionally, the iteration frequency was set to \textit{3}. Furthermore, if a higher consistency can be achieved through the prompting approach, it is advisable to increase the temperature. This can enhance the model's response variability, which positively influences the underlying CoT structure as the reasoning steps become more variable, potentially leading the model to different and better results.

However, this idea is accompanied by the greatest disadvantage of the self-consistency approach. Due to the n-fold iteration per query, costs increase both in monetary and temporal terms.

\section{Synthesis of findings and answering the research question}

This chapter consolidates the central findings of the study and synthesizes them in light of the overarching research question: To what extent can advanced prompt engineering techniques enhance the performance of LLMs on diverse sentiment analysis tasks, including binary and multi-class sentiment classification, ABSA, and irony detection? Guided by three hypotheses (H1–H3), the study systematically examined the performance of two state-of-the-art LLMs—OpenAI’s GPT-4o-mini and Google’s gemini-1.5-flash—across a variety of prompting strategies and evaluation metrics. The results present a nuanced but coherent narrative about the capabilities and limitations of prompt-based approaches in affective computing.

\subsection{H1: Prompt engineering boosts LLM accuracy in sentiment classification}

H1 is strongly supported by the experimental evidence. Across both binary (SST-2) and ternary (SB10k) classification tasks, most advanced prompting techniques led to significant improvements over the zero-shot baseline. Few-shot prompting, in particular, emerged as the most reliable strategy, substantially improving both accuracy and weighted F1-score, especially under GPT-4o-mini. This effect was especially pronounced for the neutral sentiment class in the SB10k dataset, where recall under gemini-1.5-flash improved from 0.37 to 0.51 after introducing class-balanced examples in one-shot and few-shot settings. The findings highlight that exemplar-rich prompting not only increases overall performance but also helps mitigate well-documented biases in LLMs toward polar sentiments. Importantly, while GPT-4o-mini showed the greatest relative improvement across prompting types, gemini-1.5-flash's performance gains were more constrained, indicating a difference in how model architectures respond to prompt conditioning.

\subsection{H2: Targeted prompts enhance aspect-based sentiment detection in LLMs}

H2 received only partial empirical support. Although prompting techniques generally improved results in ABSA tasks derived from the SemEval 2014 dataset, the magnitude of these gains was consistently lower compared to standard sentiment classification. This suggests that ABSA, as a more semantically complex and fine-grained task, requires more than simple prompt design to resolve sentiment toward specific aspects within contextually rich text. Notably, zero-shot-CoT and self-consistency approaches underperformed or failed to yield statistically significant improvements in this domain. These findings point to the limitations of universal prompting strategies and the need for more specialized configurations when dealing with multi-level sentiment structures. The moderate gains observed from few-shot approaches suggest that even small-scale example-based conditioning can help LLMs better handle aspectual cues, but further precision is likely required for consistent success.

\subsection{H3: Advanced prompt techniques sharpen LLM detection of ironic tweets}

H3 produced mixed results, with clear divergences across model architectures and prompting techniques. Under GPT-4o-mini, while few-shot prompting provided modest gains, both CoT and self-consistency approaches led to performance degradation. CoT-generated reasoning chains, although coherent in surface structure, frequently resulted in misclassifications, particularly when applied to ambiguous or sarcastic inputs. The reasoning steps appeared plausible yet systematically incorrect—highlighting a fundamental limitation in equating reasoning transparency with classification correctness. In stark contrast, gemini-1.5-flash exhibited substantial improvements in irony detection when CoT prompting was applied, achieving a remarkable 46\% increase in weighted F1-score compared to the baseline. This gain was especially evident in recall performance for the “no irony” class, which had previously suffered under baseline and zero-shot settings. These findings suggest that the capacity to benefit from structured reasoning is tightly coupled to the model’s internal architecture and its ability to handle subtle pragmatic cues.

\subsection{Synthesis of findings}

A set of cross-cutting insights emerges from the synthesis of these results. First, few-shot prompting consistently outperforms other approaches in terms of robustness, transferability, and ease of integration across tasks and models. Its effectiveness appears to lie in its ability to provide contextual anchors that guide the model’s internal representations during inference. Given that few-shot examples function as implicit task conditioning, these results suggest that LLMs rely heavily on exemplar-driven representation shaping, which supports recent work showing that few-shot learning operates as a form of in-context gradient descent inside transformer layers \autocite{huangTransformersLearnToImplementMultiStepGradientDescentWithChainOfThought_arXiv_2025}. Second, reasoning-based approaches like CoT and self-consistency offer mixed utility, often depending on the alignment between prompt structure and model behavior. While they have the potential to elicit richer outputs, they may also reinforce erroneous logic paths when initial assumptions are flawed. This was particularly evident in the self-consistency results under GPT-4o-mini, where multiple sampling iterations with majority voting consistently converged on confidently incorrect predictions, especially in the SST2 dataset. This highlights that reasoning chains alone do not guarantee improved classification performance; instead, their utility depends on whether the model possesses sufficiently calibrated internal heuristics to translate step-by-step reasoning into correct decision boundaries. Lastly, zero-shot-CoT approaches underperformed across nearly all tasks, despite their conceptual appeal. The systematic errors observed across zero-shot-CoT prompts further indicate that unguided reasoning amplifies model hallucination tendencies, illustrating that LLMs require externally provided semantic anchors rather than unconstrained analytical autonomy. The absence of grounding exemplars led to interpretive errors, particularly in nuanced domains like irony detection and ABSA, indicating that reasoning in a vacuum often lacks the necessary semantic context for accurate classification.

Taken together, the findings underscore that while advanced prompt engineering techniques offer substantial potential to improve LLM performance in sentiment analysis, their success is neither universal nor guaranteed. The effectiveness of a given strategy is highly contingent upon the interplay between prompt structure, task complexity, and model architecture. In particular, reasoning transparency—while intuitively desirable—does not always correlate with reasoning correctness, a distinction that becomes critical in sensitive applications such as irony detection, where the surface plausibility of a response may mask systematic misclassification.

In conclusion, the study contributes both theoretical and practical insights into how LLMs can be strategically guided through prompt engineering to perform better on affective computing tasks. The evidence presented advocates for a model-aware and task-specific approach to prompt design, in which strategies like few-shot prompting serve as a reliable foundation, while more complex techniques like CoT and self-consistency require cautious implementation and empirical validation. As LLMs continue to be deployed in real-world applications—ranging from customer sentiment monitoring to educational platforms—understanding these dynamics is crucial for ensuring both the efficacy and reliability of AI-driven sentiment analysis systems. Taken together, these insights suggest that prompt engineering should be conceptualized not as a uniform technique but as a context-dependent optimization problem shaped by semantic ambiguity, model scale, and architectural reasoning depth.

\section{Conclusion and future work}

\subsection{Key insights}

This study provides systematic empirical evidence that advanced prompt engineering techniques can significantly enhance the performance of LLMs across diverse sentiment analysis tasks. While few-shot prompting consistently yielded robust improvements in accuracy and F1-score—particularly by reducing sentiment polarity bias and improving class-level balance—reasoning-based strategies such as CoT and self-consistency exhibited notable limitations. Their effectiveness was found to be highly model-dependent, with GPT-4o-mini often producing plausible but incorrect inferences, while gemini-1.5-flash benefited substantially from CoT in complex tasks such as irony detection.

The partial confirmation of improvements in ABSA further underscores the need for task-specific prompt calibration. Across all evaluations, the divergence in performance between GPT-4o-mini and gemini-1.5-flash reaffirms that prompting strategies must be aligned with both model architecture and task complexity. This architectural sensitivity underscores that prompting strategies cannot be transferred across models without validation, as each system exhibits distinct inductive biases and differing capacities for pragmatic inference.

By uncovering the mechanisms through which LLMs interpret prompts, structure reasoning, and respond to contextual exemplars, this study contributes foundational insights to the design of more reliable, transparent, and context-aware affective computing systems. As LLMs become increasingly embedded in high-stakes domains, principled prompt design will remain essential for guiding their behavior and ensuring trustworthy deployment.

\subsection{Limitations}

This study provides empirical evidence on the efficacy of prompt engineering techniques in enhancing LLM performance on sentiment analysis and irony detection tasks. However, several limitations must be noted. First, the findings are model-specific, based solely on GPT-4o-mini and gemini-1.5-flash, limiting generalizability across architectures. Additionally, because both models are proprietary systems with opaque training corpora, it remains unclear which linguistic patterns or domain distributions may have biased their responses—an uncertainty that limits deeper causal interpretation of the results. Second, dataset sizes were downsampled to 1,000 instances per task for efficiency, which may reduce statistical power and sensitivity to rare classes. Third, prompt strategies were manually designed without adaptive tuning, potentially constraining their effectiveness. A systematic prompt ablation study could have revealed which elements of the prompt contribute most to performance gains, providing a more fine-grained understanding of prompt–model interactions. Furthermore, decoding parameters (e.g., temperature, sample size) were fixed, precluding exploration of parameter–prompt interactions. The study also lacks linguistic error analysis and does not control for multiple hypothesis testing, which may overstate significance. Furthermore, the absence of a detailed linguistic error analysis limits insight into which syntactic or semantic constructions systematically trigger misclassification, particularly in irony detection where pragmatic subtlety is crucial. Together, these factors highlight the need for broader model evaluation, automatic prompt optimization, and deeper analysis of underlying linguistic mechanisms.

\subsection{Directions for further research}

Recent advancements in LLMs have demonstrated significant potential for a range of NLP tasks, including sentiment analysis. The flexibility of prompt-based learning enables models to be rapidly adapted to new tasks without extensive fine-tuning, making prompt engineering a critical aspect of leveraging LLM capabilities. However, the effectiveness of different prompting strategies remains incompletely understood, particularly in the context of nuanced sentiment analysis scenarios. Despite promising initial results, several open questions remain regarding the alignment between specific prompting techniques and task requirements, the generalizability of findings across model architectures, and the impact of prompt design on output reliability. Addressing these questions can inform best practices for deploying LLMs in real-world sentiment analysis applications and guide future methodological and theoretical advancements in the field. Accordingly, several key research avenues emerge:

\begin{itemize}
    \item \textit{Technique efficacy across sentiment analysis tasks}: Future research could systematically evaluate whether specific prompting strategies exhibit differential efficacy across various sentiment analysis tasks. For instance, it would be valuable to investigate if few-shot prompting achieves the highest performance improvements for multi-class sentiment classification, whereas reasoning-based prompts (e.g., Chain-of-Thought prompting) yield greater benefits for more nuanced tasks such as sarcasm or irony detection.
    \item \textit{Cross-model comparisons}: An important avenue for future work is to examine how the effectiveness of prompting techniques varies across different LLM architectures. Specifically, research could explore whether optimal prompting strategies differ between different LLMs, potentially due to distinctions in their pre-training datasets, architectures, or reasoning capabilities. A more in-depth analysis that includes models of different sizes and architectures could provide valuable insights into the extent to which the effectiveness of prompt engineering techniques is model-dependent. Another promising direction is the exploration of retrieval-augmented prompting, where models receive contextual evidence before classification, potentially reducing hallucinations and improving robustness in nuanced sentiment tasks.
    \item \textit{Impact on robustness and consistency}: Further investigation is needed regarding the extent to which prompt engineering influences the robustness and consistency of model outputs. This includes assessing whether aggregation-based methods—such as self-consistency, which combines outputs from multiple reasoning trajectories—lead to more stable sentiment predictions (i.e., reduced prediction variability and error rates) relative to single-pass prompting approaches.
\end{itemize}

In summary, systematically addressing these research avenues will contribute to a deeper understanding of prompt engineering for sentiment analysis and help unlock the full potential of LLMs in this domain.

\section*{CRediT authorship contribution statement}

\textbf{Marvin Schmitt}: Conceptualization, Methodology, Software, Validation, Formal analysis, Investigation, Writing - Original Draft, Writing - Review \& Editing, Visualization. \textbf{Anne Schwerk}: Conceptualization, Writing - Original Draft, Writing - Review \& Editing, Supervision. \textbf{Sebastian Lempert}: Conceptualization, Writing - Original Draft, Writing - Review \& Editing, Supervision.

\section*{Declaration of generative AI and AI-assisted technologies in the manuscript preparation process}

During the preparation of this work, the authors utilized DeepL and GPT-4 to enhance the quality of the English language, since none of the authors are native English speakers. Furthermore, the Stanford Agentic Reviewer was used to obtain feedback on the clarity, methodology, and overall presentation of this work. The tool was used to identify potential areas for improvement and to refine the manuscript prior to submission. All revisions were carefully reviewed and approved by the authors to ensure accuracy and scholarly integrity. After using these tools, the authors reviewed and edited the content as needed and take full responsibility for the content of the published article. No generative AI tools were used to write substantive sections of the manuscript itself.

\section*{Declaration of competing interest}

The authors declare that they have no known competing financial interests or personal relationships that could have appeared to influence the work reported in this paper.

\section*{Supplementary materials}

Code is available in this GitHub-Repository: https://github.com/Marvin2108/ESCID-LLM-APET

\section*{Data availability}

Data will be made available on request.

\section*{Acknowledgements}

This work was supported by the IU International University of Applied Sciences.

\appendix

\renewcommand\thesubsection{\Alph{section}.\arabic{subsection}}

\setcounter{table}{0}
\renewcommand\thetable{A.\arabic{table}}

\setcounter{figure}{0}
\renewcommand\thefigure{A.\arabic{figure}}

\clearpage
\onecolumn
\renewcommand{\arraystretch}{1.5}
\section{All Prompts} \label{appendix:all_prompts}

\subsection{One shot prompting}

\begin{table}[!ht]
    \begin{tabularx}{\textwidth}{|C{2.5cm}|X|}
        \hline
        \textbf{Task} & \multicolumn{1}{c|}{\textbf{Prompt}} \\
        \hline
        \multirow{2}{*}{SC}
        & \textit{System}: You classify sentiments of a text. Here is an example: Input: Classify the following text into one of these two sentiments {[}’negative’, ’positive’{]}: Text: ’enriched by an imaginatively mixed cast of antic spirits’. Output: 1. \\
        \cline{2-2} 
        & \textit{User}: Classify the following text into one of these two sentiments {[}’negative’, ’positive’{]}. Text: \{text\} \\
        \hline
        \multirow{2}{*}{ABSA}
        & \textit{System}: You classify the sentiment of a specific aspect within a text. Input: Classify the sentiment for the following aspect in the following   text into one of these three sentiments {[}’negative’, ’positive’, ’neutral’{]}. Aspect: ’cord’, Text: ’I charge it at night and skip taking the cord with me because of the good battery life.’. Output: 2. \\
        \cline{2-2} 
        & \textit{User}: Classify the sentiment for the following aspect in the following text into one of these three sentiments {[}’negative’, ’positive’, ’neutral’{]}. Aspect: \{aspect\} Text: \{text\} \\
        \hline
        \multirow{2}{*}{Irony-detection}
        & \textit{System}: You determine whether a text contains ironic or sarcastic elements. Input: Does the following text contain ironic or sarcastic elements? Text: "@BaniHillal ’i feel bad for the American people, now we should invade their country and kill their children so we can save them’". Output: 1. \\
        \cline{2-2} 
        & \textit{User}: Does the following text contain ironic or sarcastic elements? Text: \{text\} \\
        \hline
\end{tabularx}%
\caption{One shot prompting}
\label{tab:prompts_one_shot}
\end{table}

\subsection{Few shot prompting}

\begin{table}[!ht]
    \begin{tabularx}{\textwidth}{|C{2.5cm}|X|}
        \hline
        \textbf{Task} & \multicolumn{1}{c|}{\textbf{Prompt}} \\
        \hline
        \multirow{2}{*}{SC}
        & \textit{System}: You classify sentiments of a text. Here are some examples: [...4 Exemplars (2 per class)...] \\
        \cline{2-2} 
        & \textit{User}: Classify the following text into one of these two sentiments [’negative’, ’positive’]. Text: \{text\} \\
        \hline
        \multirow{2}{*}{ABSA}
        & \textit{System}: You classify the sentiment of a specific aspect within a text. Here are some examples: [...2 negative, 2 positive, 3 neutral...] \\
        \cline{2-2} 
        & \textit{User}: Classify the sentiment for the following aspect in the following text into one of these three sentiments [’negative’, ’positive’, ’neutral’]. Aspect: {aspect} Text: \{text\} \\
        \hline
        \multirow{2}{*}{Irony-detection}
        & \textit{System}: You determine whether a text contains ironic or sarcastic elements. Here are some examples: [...4 Exemplars (2 per class)...] \\
        \cline{2-2} 
        & \textit{User}: Does the following text contain ironic or sarcastic elements? Text: \{text\} \\
        \hline
\end{tabularx}%
\caption{Few shot prompting}
\label{tab:prompts_few_shot}
\end{table}

\clearpage
\subsection{CoT prompting}

\begin{table}[!ht]
    \begin{tabularx}{\textwidth}{|C{2.5cm}|X|}
        \hline
        \textbf{Task} & \multicolumn{1}{c|}{\textbf{Prompt}} \\
        \hline
        \multirow{2}{*}{SC}
        & \textit{System}: You classify the sentiment of a text. Example: [...1 Exemplar with CoT for neutral class...] Output: neutral. \\
        \cline{2-2} 
        & \textit{User}: Classify the following text in one of these three sentiments [’negative’, ’positive’, ’neutral’]: Text: \{text\} \\
        \hline
        \multirow{2}{*}{ABSA}
        & \textit{System}: You classify the sentiment of a specific aspect within a text. Example: [...1 Exemplar with CoT for neutral class...] \\
        \cline{2-2} 
        & \textit{User}: Classify the sentiment for the following aspect in the following text into one of these three sentiments [’negative’, ’positive’, ’neutral’]. Aspect: {aspect} Text: \{text\} \\
        \hline
        \multirow{2}{*}{Irony-detection}
        & \textit{System}: You determine whether a text contains ironic or sarcastic elements. Example: [...1 Exemplar with CoT for positive class...] \\
        \cline{2-2} 
        & \textit{User}: Does the following text contain ironic or sarcastic elements? Text: \{text\} \\
        \hline
\end{tabularx}%
\caption{CoT prompting}
\label{tab:prompts_CoT}
\end{table}

\subsection{Zero-shot CoT prompting}

\begin{table}[!ht]
    \begin{tabularx}{\textwidth}{|C{2.5cm}|X|}
        \hline
        \textbf{Task} & \multicolumn{1}{c|}{\textbf{Prompt}} \\
        \hline
        \multirow{2}{*}{SC}
        & \textit{System}: You classify sentiments of a text. Analyze the text step-by-step to determine whether it expresses positive or negative sentiment. Explain each step in detail before providing your final answer. \\
        \cline{2-2} 
        & \textit{User}: Classify the following text into one of these two sentiments [’negative’, ’positive’]. Text: \{text\} \\
        \hline
        \multirow{2}{*}{ABSA}
        & \textit{System}: You classify the sentiment of a specific aspect within a text. Reason step-by-step by breaking down the analysis for the specified aspect. Explain each step in detail before providing your final answer. \\
        \cline{2-2} 
        & \textit{User}: Classify the sentiment for the following aspect in the following text into one of these three sentiments [’negative’, ’positive’, ’neutral’]. Aspect: {aspect} Text: \{text\} \\
        \hline
        \multirow{2}{*}{Irony-detection}
        & \textit{System}: You determine whether a text contains ironic or sarcastic elements. Analyze the text carefully and explain your reasoning step-by-step before providing the final answer. \\
        \cline{2-2} 
        & \textit{User}: Does the following text contain ironic or sarcastic elements? Text: \{text\} \\
        \hline
\end{tabularx}%
\caption{Zero-shot CoT prompting}
\label{tab:prompts_zero_shot_CoT}
\end{table}

\subsection{Self consistency prompting}

\begin{table}[!ht]
    \begin{tabularx}{\textwidth}{|C{2.5cm}|X|}
        \hline
        \textbf{Task} & \multicolumn{1}{c|}{\textbf{Prompt}} \\
        \hline
        SC
        & Application of the CoT prompt with multiple sampling rounds followed by a majority voting scheme. \\
        \hline
        ABSA
        & Application of the CoT prompt with multiple sampling rounds followed by a majority voting scheme. \\
        \hline
        Irony-detection
        & Application of the CoT prompt with multiple sampling rounds followed by a majority voting scheme. \\
        \hline
\end{tabularx}%
\caption{Self consistency prompting}
\label{tab:prompts_self_consistency}
\end{table}

\setcounter{table}{0}
\renewcommand\thetable{B.\arabic{table}}

\setcounter{figure}{0}
\renewcommand\thefigure{B.\arabic{figure}}

\clearpage
\section{Statistical testing (Bootstrap)}\label{appendix:statTest}

This section presents the results of the statistical significance analysis based on bootstrap resampling. Instead of relying on p-values, statistical significance is assessed via a 95 \% bootstrap confidence interval of the F1-score difference. If the interval does not contain zero, the difference is considered statistically significant. 

\begin{table}[!ht]
\resizebox{\textwidth}{!}{%
\begin{tabular}{l|cccc|cccc|}
\cline{2-9}
                                                & \multicolumn{4}{c|}{\textbf{SC (SST2)}}                                                                                                                                                              & \multicolumn{4}{c|}{\textbf{SC (SB10k)}}                                                                                                                                                              \\ \hline
\multicolumn{1}{|l|}{\textbf{LLM}}              & \multicolumn{2}{c|}{GPT}                                                                                  & \multicolumn{2}{c|}{Gemini}                                                         & \multicolumn{2}{c|}{GPT}                                                                                  & \multicolumn{2}{c|}{Gemini}                                                          \\ \hline
	\multicolumn{1}{|l|}{\textbf{PE-Approach}}
&	\multicolumn{1}{C{3cm}|}{95\% confidence interval of the F1 difference}
&	\multicolumn{1}{C{3cm}|}{Significance}
&	\multicolumn{1}{C{3cm}|}{95\% confidence interval of the F1 difference}
&	\multicolumn{1}{C{3cm}|}{Significance}
&	\multicolumn{1}{C{3cm}|}{95\% confidence interval of the F1 difference}
&	\multicolumn{1}{C{3cm}|}{Significance}
&	\multicolumn{1}{C{3cm}|}{95\% confidence interval of the F1 difference}
&	\multicolumn{1}{C{3cm}|}{Significance}  \\ \hline
\multicolumn{1}{|l|}{\textbf{One-Shot}}         & \multicolumn{1}{c|}{{[}-0.0288, -0.0004{]}}                          & \multicolumn{1}{c|}{base}          & \multicolumn{1}{c|}{{[}0.0389, 0.0751{]}}                            & PE           & \multicolumn{1}{c|}{{[}0.0288,   0.0684{]}}                          & \multicolumn{1}{c|}{PE}            & \multicolumn{1}{c|}{{[}0.0540,   0.1004{]}}                          & PE            \\ \hline
\multicolumn{1}{|l|}{\textbf{Few-Shot}}         & \multicolumn{1}{c|}{{[}0.0673, 0.1195{]}}                            & \multicolumn{1}{c|}{PE}            & \multicolumn{1}{c|}{{[}0.0549, 0.0931{]}}                            & PE           & \multicolumn{1}{c|}{{[}0.0673, 0.1195{]}}                            & \multicolumn{1}{c|}{PE}            & \multicolumn{1}{c|}{{[}0.0549, 0.1005{]}}                            & PE            \\ \hline
\multicolumn{1}{|l|}{\textbf{CoT}}              & \multicolumn{1}{c|}{{[}-0.0197, 0.0134{]}}                           & \multicolumn{1}{c|}{No difference} & \multicolumn{1}{c|}{{[}0.0397, 0.0801{]}}                            & PE           & \multicolumn{1}{c|}{{[}-0.0267, 0.0239{]}}                           & \multicolumn{1}{c|}{No difference} & \multicolumn{1}{c|}{{[}-0.0082, 0.0483{]}}                           & No difference \\ \hline
\multicolumn{1}{|l|}{\textbf{Zero-Shot CoT}}    & \multicolumn{1}{c|}{{[}-0.0055, 0.0233{]}}                           & \multicolumn{1}{c|}{No difference} & \multicolumn{1}{c|}{{[}0.0002, 0.0462{]}}                            & PE           & \multicolumn{1}{c|}{{[}-0.1895, -0.121{]}}                           & \multicolumn{1}{c|}{base}          & \multicolumn{1}{c|}{{[}-0.0403, 0.0145{]}}                           & No difference \\ \hline
\multicolumn{1}{|l|}{\textbf{Self-Consistency}} & \multicolumn{1}{c|}{{[}-0.1053, -0.0597{]}}                          & \multicolumn{1}{c|}{base}          & \multicolumn{1}{c|}{{[}0.0278, 0.0698{]}}                            & PE           & \multicolumn{1}{c|}{{[}-0.0114, 0.0373{]}}                           & \multicolumn{1}{c|}{No difference} & \multicolumn{1}{c|}{{[}0.0201, 0.0736{]}}                            & PE            \\ \hline
\end{tabular}%
}
\caption{Bootstrap sentiment classification (base = baseline approach delivers significantly better results, PE = prompt approach delivers significantly better results, No difference = no statistically significant difference was found between the results)}
\label{tab:BootstrapSC}
\end{table}

\begin{table}[!ht]
\resizebox{\textwidth}{!}{%
\begin{tabular}{l|cccc|cccc|}
\cline{2-9}
                                                & \multicolumn{4}{c|}{\textbf{ABSA (SemEval-2014)}}                                                                                                                                                              & \multicolumn{4}{c|}{\textbf{Irony-detection (SemEval-2018)}}                                                                                                                                                              \\ \hline
\multicolumn{1}{|l|}{\textbf{LLM}}              & \multicolumn{2}{c|}{GPT}                                                                                  & \multicolumn{2}{c|}{Gemini}                                                         & \multicolumn{2}{c|}{GPT}                                                                                  & \multicolumn{2}{c|}{Gemini}                                                          \\ \hline
	\multicolumn{1}{|l|}{\textbf{PE-Approach}}
&	\multicolumn{1}{C{3cm}|}{95\% confidence interval of the F1 difference}
&	\multicolumn{1}{C{3cm}|}{Significance}
&	\multicolumn{1}{C{3cm}|}{95\% confidence interval of the F1 difference}
&	\multicolumn{1}{C{3cm}|}{Significance}
&	\multicolumn{1}{C{3cm}|}{95\% confidence interval of the F1 difference}
&	\multicolumn{1}{C{3cm}|}{Significance}
&	\multicolumn{1}{C{3cm}|}{95\% confidence interval of the F1 difference}
&	\multicolumn{1}{C{3cm}|}{Significance}  \\ \hline
\multicolumn{1}{|l|}{\textbf{One-Shot}}         & \multicolumn{1}{c|}{{[}0.0003, 0.0267{]}}                            & \multicolumn{1}{c|}{PE}            & \multicolumn{1}{c|}{{[}-0.0143, 0.0169{]}}                           & No difference & \multicolumn{1}{c|}{{[}-0.0331,   0.0082{]}}                         & \multicolumn{1}{c|}{No difference} & \multicolumn{1}{c|}{{[}0.1318,   0.1931{]}}                          & PE           \\ \hline
\multicolumn{1}{|l|}{\textbf{Few-Shot}}         & \multicolumn{1}{c|}{{[}0.0053, 0.0364{]}}                            & \multicolumn{1}{c|}{PE}            & \multicolumn{1}{c|}{{[}-0.0043, 0.0335{]}}                           & No difference & \multicolumn{1}{c|}{{[}-0.0024, 0.0404{]}}                           & \multicolumn{1}{c|}{No difference} & \multicolumn{1}{c|}{{[}0.1248, 0.1877{]}}                            & PE           \\ \hline
\multicolumn{1}{|l|}{\textbf{CoT}}              & \multicolumn{1}{c|}{{[}-0.0220, 0.0111{]}}                           & \multicolumn{1}{c|}{No difference} & \multicolumn{1}{c|}{{[}-0.0055, 0.0352{]}}                           & No difference & \multicolumn{1}{c|}{{[}-0.1091, -0.0531{]}}                          & \multicolumn{1}{c|}{base}          & \multicolumn{1}{c|}{{[}0.1578, 0.2282{]}}                            & PE           \\ \hline
\multicolumn{1}{|l|}{\textbf{Zero-Shot CoT}}    & \multicolumn{1}{c|}{{[}-0.0460, -0.0051{]}}                          & \multicolumn{1}{c|}{base}          & \multicolumn{1}{c|}{{[}-0.0422, -0.0025{]}}                          & base          & \multicolumn{1}{c|}{{[}-0.1504, -0.0868{]}}                          & \multicolumn{1}{c|}{base}          & \multicolumn{1}{c|}{{[}0.0155, 0.0599{]}}                            & PE           \\ \hline
\multicolumn{1}{|l|}{\textbf{Self-Consistency}} & \multicolumn{1}{c|}{{[}-0.0254, 0.0066{]}}                           & \multicolumn{1}{c|}{No difference} & \multicolumn{1}{c|}{{[}-0.0039, 0.0373{]}}                           & No difference & \multicolumn{1}{c|}{{[}-0.2014, -0.1265{]}}                          & \multicolumn{1}{c|}{base}          & \multicolumn{1}{c|}{{[}0.0317, 0.0799{]}}                            & PE           \\ \hline
\end{tabular}%
}
\caption{Bootstrap ABSA \& Irony (base = baseline approach delivers significantly better results, PE = prompt approach delivers significantly better results, No difference = no statistically significant difference was found between the results)}
\label{tab:BootstrapABSA}
\end{table}

\setcounter{table}{0}
\renewcommand\thetable{C.\arabic{table}}

\setcounter{figure}{0}
\renewcommand\thefigure{C.\arabic{figure}}

\clearpage
\section{All results}\label{appendix:all_results}

\subsection{Results sentiment-classification} \label{appendix_section_results_sentiment_classification}

\begin{table}[!ht]
\resizebox{\textwidth}{!}{%
\begin{tabular}{cl|cccccc|cccccccc|}
\cline{3-16}
\multicolumn{1}{l}{\textbf{}}                           &                                      & \multicolumn{6}{c|}{\textbf{SST2}}                                                                                                                                                                                                            & \multicolumn{8}{c|}{\textbf{SB10k}}                                                                                                                                                                                                                                                                                           \\ \cline{3-16} 
\multicolumn{1}{l}{\textbf{}}                           &                                      & \multicolumn{3}{c|}{\textbf{GPT}}                                                                                     & \multicolumn{3}{c|}{\textbf{GEMINI}}                                                                                  & \multicolumn{4}{c|}{\textbf{GPT}}                                                                                                                             & \multicolumn{4}{c|}{\textbf{GEMINI}}                                                                                                                          \\ \hline
\multicolumn{1}{|c|}{\textbf{PE Approach}}              & \multicolumn{1}{c|}{\textbf{Metric}} & \multicolumn{1}{l|}{\textbf{Negative}} & \multicolumn{1}{l|}{\textbf{Positive}} & \multicolumn{1}{l|}{\textbf{Total}} & \multicolumn{1}{l|}{\textbf{Negative}} & \multicolumn{1}{l|}{\textbf{Positive}} & \multicolumn{1}{l|}{\textbf{Total}} & \multicolumn{1}{l|}{\textbf{Negative}} & \multicolumn{1}{l|}{\textbf{Positive}} & \multicolumn{1}{l|}{\textbf{Neutral}} & \multicolumn{1}{l|}{\textbf{Total}} & \multicolumn{1}{l|}{\textbf{Negative}} & \multicolumn{1}{l|}{\textbf{Positive}} & \multicolumn{1}{l|}{\textbf{Neutral}} & \multicolumn{1}{l|}{\textbf{Total}} \\ \hline
\multicolumn{1}{|c|}{\multirow{5}{*}{BASE}}             & Precision                            & \multicolumn{1}{c|}{0,85}              & \multicolumn{1}{c|}{0,97}              & \multicolumn{1}{c|}{0,91}           & \multicolumn{1}{c|}{0,75}              & \multicolumn{1}{c|}{0,99}              & 0,87                                & \multicolumn{1}{c|}{0,41}              & \multicolumn{1}{c|}{0,58}              & \multicolumn{1}{c|}{0,84}             & \multicolumn{1}{c|}{0,61}           & \multicolumn{1}{c|}{0,33}              & \multicolumn{1}{c|}{0,55}              & \multicolumn{1}{c|}{0,83}             & 0,57                                \\ \cline{2-16} 
\multicolumn{1}{|c|}{}                                  & Recall                               & \multicolumn{1}{c|}{0,97}              & \multicolumn{1}{c|}{0,86}              & \multicolumn{1}{c|}{0,92}           & \multicolumn{1}{c|}{0,99}              & \multicolumn{1}{c|}{0,72}              & 0,86                                & \multicolumn{1}{c|}{0,87}              & \multicolumn{1}{c|}{0,72}              & \multicolumn{1}{c|}{0,52}             & \multicolumn{1}{c|}{0,70}           & \multicolumn{1}{c|}{0,94}              & \multicolumn{1}{c|}{0,65}              & \multicolumn{1}{c|}{0,37}             & 0,65                                \\ \cline{2-16} 
\multicolumn{1}{|c|}{}                                  & F1 (macro)                           & \multicolumn{1}{c|}{0,91}              & \multicolumn{1}{c|}{0,91}              & \multicolumn{1}{c|}{0,91}           & \multicolumn{1}{c|}{0,86}              & \multicolumn{1}{c|}{0,84}              & 0,85                                & \multicolumn{1}{c|}{0,56}              & \multicolumn{1}{c|}{0,64}              & \multicolumn{1}{c|}{0,64}             & \multicolumn{1}{c|}{0,61}           & \multicolumn{1}{c|}{0,49}              & \multicolumn{1}{c|}{0,59}              & \multicolumn{1}{c|}{0,51}             & 0,53                                \\ \cline{2-16} 
\multicolumn{1}{|c|}{}                                  & Weighted Avg F1                      & \multicolumn{1}{c|}{}                  & \multicolumn{1}{c|}{}                  & \multicolumn{1}{c|}{0,91}           & \multicolumn{1}{c|}{}                  & \multicolumn{1}{c|}{}                  & 0,85                                & \multicolumn{1}{c|}{}                  & \multicolumn{1}{c|}{}                  & \multicolumn{1}{c|}{}                 & \multicolumn{1}{c|}{0,63}           & \multicolumn{1}{c|}{}                  & \multicolumn{1}{c|}{}                  & \multicolumn{1}{c|}{}                 & 0,53                                \\ \cline{2-16} 
\multicolumn{1}{|c|}{}                                  & Accuracy                             & \multicolumn{1}{c|}{}                  & \multicolumn{1}{c|}{}                  & \multicolumn{1}{c|}{0,91}           & \multicolumn{1}{c|}{}                  & \multicolumn{1}{c|}{}                  & 0,85                                & \multicolumn{1}{c|}{}                  & \multicolumn{1}{c|}{}                  & \multicolumn{1}{c|}{}                 & \multicolumn{1}{c|}{0,62}           & \multicolumn{1}{c|}{}                  & \multicolumn{1}{c|}{}                  & \multicolumn{1}{c|}{}                 & 0,52                                \\ \hline
\multicolumn{1}{|c|}{\multirow{5}{*}{One Shot}}         & Precision                            & \multicolumn{1}{c|}{0,84}              & \multicolumn{1}{c|}{0,99}              & \multicolumn{1}{c|}{0,92}           & \multicolumn{1}{c|}{0,81}              & \multicolumn{1}{c|}{0,99}              & 0,90                                & \multicolumn{1}{c|}{0,45}              & \multicolumn{1}{c|}{0,62}              & \multicolumn{1}{c|}{0,86}             & \multicolumn{1}{c|}{0,64}           & \multicolumn{1}{c|}{0,36}              & \multicolumn{1}{c|}{0,61}              & \multicolumn{1}{c|}{0,81}             & 0,59                                \\ \cline{2-16} 
\multicolumn{1}{|c|}{}                                  & Recall                               & \multicolumn{1}{c|}{0,99}              & \multicolumn{1}{c|}{0,84}              & \multicolumn{1}{c|}{0,92}           & \multicolumn{1}{c|}{0,99}              & \multicolumn{1}{c|}{0,81}              & 0,90                                & \multicolumn{1}{c|}{0,88}              & \multicolumn{1}{c|}{0,7}               & \multicolumn{1}{c|}{0,6}              & \multicolumn{1}{c|}{0,73}           & \multicolumn{1}{c|}{0,88}              & \multicolumn{1}{c|}{0,62}              & \multicolumn{1}{c|}{0,51}             & 0,67                                \\ \cline{2-16} 
\multicolumn{1}{|c|}{}                                  & F1 (macro)                           & \multicolumn{1}{c|}{0,91}              & \multicolumn{1}{c|}{0,91}              & \multicolumn{1}{c|}{0,91}           & \multicolumn{1}{c|}{0,89}              & \multicolumn{1}{c|}{0,89}              & 0,89                                & \multicolumn{1}{c|}{0,6}               & \multicolumn{1}{c|}{0,66}              & \multicolumn{1}{c|}{0,7}              & \multicolumn{1}{c|}{0,65}           & \multicolumn{1}{c|}{0,52}              & \multicolumn{1}{c|}{0,62}              & \multicolumn{1}{c|}{0,62}             & 0,59                                \\ \cline{2-16} 
\multicolumn{1}{|c|}{}                                  & Weighted Avg F1                      & \multicolumn{1}{c|}{}                  & \multicolumn{1}{c|}{}                  & \multicolumn{1}{c|}{0,91}           & \multicolumn{1}{c|}{}                  & \multicolumn{1}{c|}{}                  & 0,89                                & \multicolumn{1}{c|}{}                  & \multicolumn{1}{c|}{}                  & \multicolumn{1}{c|}{}                 & \multicolumn{1}{c|}{0,68}           & \multicolumn{1}{c|}{}                  & \multicolumn{1}{c|}{}                  & \multicolumn{1}{c|}{}                 & 0,60                                \\ \cline{2-16} 
\multicolumn{1}{|c|}{}                                  & Accuracy                             & \multicolumn{1}{c|}{}                  & \multicolumn{1}{c|}{}                  & \multicolumn{1}{c|}{0,91}           & \multicolumn{1}{c|}{}                  & \multicolumn{1}{c|}{}                  & 0,89                                & \multicolumn{1}{c|}{}                  & \multicolumn{1}{c|}{}                  & \multicolumn{1}{c|}{}                 & \multicolumn{1}{c|}{0,67}           & \multicolumn{1}{c|}{}                  & \multicolumn{1}{c|}{}                  & \multicolumn{1}{c|}{}                 & 0,59                                \\ \hline
\multicolumn{1}{|c|}{\multirow{5}{*}{Few Shot}}         & Precision                            & \multicolumn{1}{c|}{0,88}              & \multicolumn{1}{c|}{0,98}              & \multicolumn{1}{c|}{0,93}           & \multicolumn{1}{c|}{0,84}              & \multicolumn{1}{c|}{0,99}              & 0,92                                & \multicolumn{1}{c|}{0,52}              & \multicolumn{1}{c|}{0,71}              & \multicolumn{1}{c|}{0,81}             & \multicolumn{1}{c|}{0,68}           & \multicolumn{1}{c|}{0,39}              & \multicolumn{1}{c|}{0,58}              & \multicolumn{1}{c|}{0,82}             & 0,60                                \\ \cline{2-16} 
\multicolumn{1}{|c|}{}                                  & Recall                               & \multicolumn{1}{c|}{0,98}              & \multicolumn{1}{c|}{0,89}              & \multicolumn{1}{c|}{0,94}           & \multicolumn{1}{c|}{0,99}              & \multicolumn{1}{c|}{0,84}              & 0,92                                & \multicolumn{1}{c|}{0,82}              & \multicolumn{1}{c|}{0,6}               & \multicolumn{1}{c|}{0,74}             & \multicolumn{1}{c|}{0,72}           & \multicolumn{1}{c|}{0,89}              & \multicolumn{1}{c|}{0,66}              & \multicolumn{1}{c|}{0,5}              & 0,68                                \\ \cline{2-16} 
\multicolumn{1}{|c|}{}                                  & F1 (macro)                           & \multicolumn{1}{c|}{0,93}              & \multicolumn{1}{c|}{0,93}              & \multicolumn{1}{c|}{0,93}           & \multicolumn{1}{c|}{0,91}              & \multicolumn{1}{c|}{0,91}              & 0,91                                & \multicolumn{1}{c|}{0,64}              & \multicolumn{1}{c|}{0,65}              & \multicolumn{1}{c|}{0,77}             & \multicolumn{1}{c|}{0,69}           & \multicolumn{1}{c|}{0,54}              & \multicolumn{1}{c|}{0,62}              & \multicolumn{1}{c|}{0,62}             & 0,59                                \\ \cline{2-16} 
\multicolumn{1}{|c|}{}                                  & Weighted Avg F1                      & \multicolumn{1}{c|}{}                  & \multicolumn{1}{c|}{}                  & \multicolumn{1}{c|}{0,93}           & \multicolumn{1}{c|}{}                  & \multicolumn{1}{c|}{}                  & 0,91                                & \multicolumn{1}{c|}{}                  & \multicolumn{1}{c|}{}                  & \multicolumn{1}{c|}{}                 & \multicolumn{1}{c|}{0,72}           & \multicolumn{1}{c|}{}                  & \multicolumn{1}{c|}{}                  & \multicolumn{1}{c|}{}                 & 0,61                                \\ \cline{2-16} 
\multicolumn{1}{|c|}{}                                  & Accuracy                             & \multicolumn{1}{c|}{}                  & \multicolumn{1}{c|}{}                  & \multicolumn{1}{c|}{0,93}           & \multicolumn{1}{c|}{}                  & \multicolumn{1}{c|}{}                  & 0,91                                & \multicolumn{1}{c|}{}                  & \multicolumn{1}{c|}{}                  & \multicolumn{1}{c|}{}                 & \multicolumn{1}{c|}{0,72}           & \multicolumn{1}{c|}{}                  & \multicolumn{1}{c|}{}                  & \multicolumn{1}{c|}{}                 & 0,6                                 \\ \hline
\multicolumn{1}{|c|}{\multirow{5}{*}{CoT}}              & Precision                            & \multicolumn{1}{c|}{0,87}              & \multicolumn{1}{c|}{0,99}              & \multicolumn{1}{c|}{0,93}           & \multicolumn{1}{c|}{0,94}              & \multicolumn{1}{c|}{0,97}              & 0,96                                & \multicolumn{1}{c|}{0,42}              & \multicolumn{1}{c|}{0,56}              & \multicolumn{1}{c|}{0,88}             & \multicolumn{1}{c|}{0,62}           & \multicolumn{1}{c|}{0,38}              & \multicolumn{1}{c|}{0,51}              & \multicolumn{1}{c|}{0,9}              & 0,60                                \\ \cline{2-16} 
\multicolumn{1}{|c|}{}                                  & Recall                               & \multicolumn{1}{c|}{0,99}              & \multicolumn{1}{c|}{0,87}              & \multicolumn{1}{c|}{0,93}           & \multicolumn{1}{c|}{0,97}              & \multicolumn{1}{c|}{0,94}              & 0,96                                & \multicolumn{1}{c|}{0,87}              & \multicolumn{1}{c|}{0,78}              & \multicolumn{1}{c|}{0,49}             & \multicolumn{1}{c|}{0,71}           & \multicolumn{1}{c|}{0,91}              & \multicolumn{1}{c|}{0,79}              & \multicolumn{1}{c|}{0,36}             & 0,69                                \\ \cline{2-16} 
\multicolumn{1}{|c|}{}                                  & F1 (macro)                           & \multicolumn{1}{c|}{0,93}              & \multicolumn{1}{c|}{0,92}              & \multicolumn{1}{c|}{0,93}           & \multicolumn{1}{c|}{0,95}              & \multicolumn{1}{c|}{0,96}              & 0,96                                & \multicolumn{1}{c|}{0,56}              & \multicolumn{1}{c|}{0,65}              & \multicolumn{1}{c|}{0,63}             & \multicolumn{1}{c|}{0,61}           & \multicolumn{1}{c|}{0,53}              & \multicolumn{1}{c|}{0,62}              & \multicolumn{1}{c|}{0,52}             & 0,56                                \\ \cline{2-16} 
\multicolumn{1}{|c|}{}                                  & Weighted Avg F1                      & \multicolumn{1}{c|}{}                  & \multicolumn{1}{c|}{}                  & \multicolumn{1}{c|}{0,92}           & \multicolumn{1}{c|}{}                  & \multicolumn{1}{c|}{}                  & 0,95                                & \multicolumn{1}{c|}{}                  & \multicolumn{1}{c|}{}                  & \multicolumn{1}{c|}{}                 & \multicolumn{1}{c|}{0,63}           & \multicolumn{1}{c|}{}                  & \multicolumn{1}{c|}{}                  & \multicolumn{1}{c|}{}                 & 0,54                                \\ \cline{2-16} 
\multicolumn{1}{|c|}{}                                  & Accuracy                             & \multicolumn{1}{c|}{}                  & \multicolumn{1}{c|}{}                  & \multicolumn{1}{c|}{0,92}           & \multicolumn{1}{c|}{}                  & \multicolumn{1}{c|}{}                  & 0,95                                & \multicolumn{1}{c|}{}                  & \multicolumn{1}{c|}{}                  & \multicolumn{1}{c|}{}                 & \multicolumn{1}{c|}{0,62}           & \multicolumn{1}{c|}{}                  & \multicolumn{1}{c|}{}                  & \multicolumn{1}{c|}{}                 & 0,55                                \\ \hline
\multicolumn{1}{|c|}{\multirow{5}{*}{ZeroShot-CoT}}     & Precision                            & \multicolumn{1}{c|}{0,87}              & \multicolumn{1}{c|}{0,97}              & \multicolumn{1}{c|}{0,92}           & \multicolumn{1}{c|}{0,81}              & \multicolumn{1}{c|}{0,96}              & 0,89                                & \multicolumn{1}{c|}{0,37}              & \multicolumn{1}{c|}{0,46}              & \multicolumn{1}{c|}{0,83}             & \multicolumn{1}{c|}{0,55}           & \multicolumn{1}{c|}{0,32}              & \multicolumn{1}{c|}{0,52}              & \multicolumn{1}{c|}{0,8}              & 0,55                                \\ \cline{2-16} 
\multicolumn{1}{|c|}{}                                  & Recall                               & \multicolumn{1}{c|}{0,97}              & \multicolumn{1}{c|}{0,88}              & \multicolumn{1}{c|}{0,93}           & \multicolumn{1}{c|}{0,96}              & \multicolumn{1}{c|}{0,8}               & 0,88                                & \multicolumn{1}{c|}{0,91}              & \multicolumn{1}{c|}{0,78}              & \multicolumn{1}{c|}{0,28}             & \multicolumn{1}{c|}{0,66}           & \multicolumn{1}{c|}{0,93}              & \multicolumn{1}{c|}{0,6}               & \multicolumn{1}{c|}{0,36}             & 0,63                                \\ \cline{2-16} 
\multicolumn{1}{|c|}{}                                  & F1 (macro)                           & \multicolumn{1}{c|}{0,92}              & \multicolumn{1}{c|}{0,92}              & \multicolumn{1}{c|}{0,92}           & \multicolumn{1}{c|}{0,88}              & \multicolumn{1}{c|}{0,87}              & 0,88                                & \multicolumn{1}{c|}{0,52}              & \multicolumn{1}{c|}{0,58}              & \multicolumn{1}{c|}{0,42}             & \multicolumn{1}{c|}{0,51}           & \multicolumn{1}{c|}{0,48}              & \multicolumn{1}{c|}{0,56}              & \multicolumn{1}{c|}{0,5}              & 0,51                                \\ \cline{2-16} 
\multicolumn{1}{|c|}{}                                  & Weighted Avg F1                      & \multicolumn{1}{c|}{}                  & \multicolumn{1}{c|}{}                  & \multicolumn{1}{c|}{0,92}           & \multicolumn{1}{c|}{}                  & \multicolumn{1}{c|}{}                  & 0,87                                & \multicolumn{1}{c|}{}                  & \multicolumn{1}{c|}{}                  & \multicolumn{1}{c|}{}                 & \multicolumn{1}{c|}{0,47}           & \multicolumn{1}{c|}{}                  & \multicolumn{1}{c|}{}                  & \multicolumn{1}{c|}{}                 & 0,51                                \\ \cline{2-16} 
\multicolumn{1}{|c|}{}                                  & Accuracy                             & \multicolumn{1}{c|}{}                  & \multicolumn{1}{c|}{}                  & \multicolumn{1}{c|}{0,92}           & \multicolumn{1}{c|}{}                  & \multicolumn{1}{c|}{}                  & 0,87                                & \multicolumn{1}{c|}{}                  & \multicolumn{1}{c|}{}                  & \multicolumn{1}{c|}{}                 & \multicolumn{1}{c|}{0,5}            & \multicolumn{1}{c|}{}                  & \multicolumn{1}{c|}{}                  & \multicolumn{1}{c|}{}                 & 0,51                                \\ \hline
\multicolumn{1}{|c|}{\multirow{5}{*}{Self-Consistency}} & Precision                            & \multicolumn{1}{c|}{0,76}              & \multicolumn{1}{c|}{0,99}              & \multicolumn{1}{c|}{0,88}           & \multicolumn{1}{c|}{0,82}              & \multicolumn{1}{c|}{0,97}              & 0,90                                & \multicolumn{1}{c|}{0,42}              & \multicolumn{1}{c|}{0,57}              & \multicolumn{1}{c|}{0,57}             & \multicolumn{1}{c|}{0,52}           & \multicolumn{1}{c|}{0,39}              & \multicolumn{1}{c|}{0,53}              & \multicolumn{1}{c|}{0,9}              & 0,61                                \\ \cline{2-16} 
\multicolumn{1}{|c|}{}                                  & Recall                               & \multicolumn{1}{c|}{0,99}              & \multicolumn{1}{c|}{0,73}              & \multicolumn{1}{c|}{0,86}           & \multicolumn{1}{c|}{0,97}              & \multicolumn{1}{c|}{0,82}              & 0,90                                & \multicolumn{1}{c|}{0,89}              & \multicolumn{1}{c|}{0,76}              & \multicolumn{1}{c|}{0,65}             & \multicolumn{1}{c|}{0,77}           & \multicolumn{1}{c|}{0,93}              & \multicolumn{1}{c|}{0,79}              & \multicolumn{1}{c|}{0,41}             & 0,71                                \\ \cline{2-16} 
\multicolumn{1}{|c|}{}                                  & F1 (macro)                           & \multicolumn{1}{c|}{0,86}              & \multicolumn{1}{c|}{0,84}              & \multicolumn{1}{c|}{0,85}           & \multicolumn{1}{c|}{0,89}              & \multicolumn{1}{c|}{0,89}              & 0,89                                & \multicolumn{1}{c|}{0,57}              & \multicolumn{1}{c|}{0,65}              & \multicolumn{1}{c|}{0,65}             & \multicolumn{1}{c|}{0,62}           & \multicolumn{1}{c|}{0,55}              & \multicolumn{1}{c|}{0,63}              & \multicolumn{1}{c|}{0,56}             & 0,58                                \\ \cline{2-16} 
\multicolumn{1}{|c|}{}                                  & Weighted Avg F1                      & \multicolumn{1}{c|}{}                  & \multicolumn{1}{c|}{}                  & \multicolumn{1}{c|}{0,85}           & \multicolumn{1}{c|}{}                  & \multicolumn{1}{c|}{}                  & 0,89                                & \multicolumn{1}{c|}{}                  & \multicolumn{1}{c|}{}                  & \multicolumn{1}{c|}{}                 & \multicolumn{1}{c|}{0,64}           & \multicolumn{1}{c|}{}                  & \multicolumn{1}{c|}{}                  & \multicolumn{1}{c|}{}                 & 0,58                                \\ \cline{2-16} 
\multicolumn{1}{|c|}{}                                  & Accuracy                             & \multicolumn{1}{c|}{}                  & \multicolumn{1}{c|}{}                  & \multicolumn{1}{c|}{0,85}           & \multicolumn{1}{c|}{}                  & \multicolumn{1}{c|}{}                  & 0,89                                & \multicolumn{1}{c|}{}                  & \multicolumn{1}{c|}{}                  & \multicolumn{1}{c|}{}                 & \multicolumn{1}{c|}{0,63}           & \multicolumn{1}{c|}{}                  & \multicolumn{1}{c|}{}                  & \multicolumn{1}{c|}{}                 & 0,58                                \\ \hline
\end{tabular}%
}
\caption{All results sentiment-classification}
\label{tab:Result_sc}
\end{table}

\clearpage
\subsection{Results aspect-based sentiment analysis} \label{appendix_section_results_ABSA}

\begin{table}[!ht]
\resizebox{\textwidth}{!}{%
\begin{tabular}{cl|cccccccc|}
\cline{3-10}
\multicolumn{1}{l}{}                                    &                                      & \multicolumn{8}{c|}{\textbf{SemEval-2014}}                                                                                                                                                                                                                                                                                    \\ \cline{3-10} 
\multicolumn{1}{l}{}                                    &                                      & \multicolumn{4}{c|}{\textbf{GPT}}                                                                                                                             & \multicolumn{4}{c|}{\textbf{GEMINI}}                                                                                                                          \\ \hline
\multicolumn{1}{|c|}{\textbf{PE Approach}}              & \multicolumn{1}{c|}{\textbf{Metric}} & \multicolumn{1}{l|}{\textbf{Negative}} & \multicolumn{1}{l|}{\textbf{Positive}} & \multicolumn{1}{l|}{\textbf{Neutral}} & \multicolumn{1}{l|}{\textbf{Total}} & \multicolumn{1}{l|}{\textbf{Negative}} & \multicolumn{1}{l|}{\textbf{Positive}} & \multicolumn{1}{l|}{\textbf{Neutral}} & \multicolumn{1}{l|}{\textbf{Total}} \\ \hline
\multicolumn{1}{|c|}{\multirow{5}{*}{BASE}}             & Precision                            & \multicolumn{1}{c|}{0,84}              & \multicolumn{1}{c|}{0,87}              & \multicolumn{1}{c|}{0,7}              & \multicolumn{1}{c|}{0,80}           & \multicolumn{1}{c|}{0,79}              & \multicolumn{1}{c|}{0,88}              & \multicolumn{1}{c|}{0,75}             & 0,81                                \\ \cline{2-10} 
\multicolumn{1}{|c|}{}                                  & Recall                               & \multicolumn{1}{c|}{0,91}              & \multicolumn{1}{c|}{0,92}              & \multicolumn{1}{c|}{0,46}             & \multicolumn{1}{c|}{0,76}           & \multicolumn{1}{c|}{0,95}              & \multicolumn{1}{c|}{0,92}              & \multicolumn{1}{c|}{0,34}             & 0,63                                \\ \cline{2-10} 
\multicolumn{1}{|c|}{}                                  & F1 (macro)                           & \multicolumn{1}{c|}{0,87}              & \multicolumn{1}{c|}{0,89}              & \multicolumn{1}{c|}{0,56}             & \multicolumn{1}{c|}{0,77}           & \multicolumn{1}{c|}{0,86}              & \multicolumn{1}{c|}{0,9}               & \multicolumn{1}{c|}{0,46}             & 0,68                                \\ \cline{2-10} 
\multicolumn{1}{|c|}{}                                  & Weighted Avg F1                      & \multicolumn{1}{c|}{}                  & \multicolumn{1}{c|}{}                  & \multicolumn{1}{c|}{}                 & \multicolumn{1}{c|}{0,83}           & \multicolumn{1}{c|}{}                  & \multicolumn{1}{c|}{}                  & \multicolumn{1}{c|}{}                 & 0,81                                \\ \cline{2-10} 
\multicolumn{1}{|c|}{}                                  & Accuracy                             & \multicolumn{1}{c|}{}                  & \multicolumn{1}{c|}{}                  & \multicolumn{1}{c|}{}                 & \multicolumn{1}{c|}{0,84}           & \multicolumn{1}{c|}{}                  & \multicolumn{1}{c|}{}                  & \multicolumn{1}{c|}{}                 & 0,83                                \\ \hline
\multicolumn{1}{|c|}{\multirow{5}{*}{One Shot}}         & Precision                            & \multicolumn{1}{c|}{0,84}              & \multicolumn{1}{c|}{0,88}              & \multicolumn{1}{c|}{0,72}             & \multicolumn{1}{c|}{0,81}           & \multicolumn{1}{c|}{0,79}              & \multicolumn{1}{c|}{0,88}              & \multicolumn{1}{c|}{0,74}             & 0,80                                \\ \cline{2-10} 
\multicolumn{1}{|c|}{}                                  & Recall                               & \multicolumn{1}{c|}{0,9}               & \multicolumn{1}{c|}{0,92}              & \multicolumn{1}{c|}{0,53}             & \multicolumn{1}{c|}{0,78}           & \multicolumn{1}{c|}{0,94}              & \multicolumn{1}{c|}{0,91}              & \multicolumn{1}{c|}{0,37}             & 0,74                                \\ \cline{2-10} 
\multicolumn{1}{|c|}{}                                  & F1 (macro)                           & \multicolumn{1}{c|}{0,87}              & \multicolumn{1}{c|}{0,9}               & \multicolumn{1}{c|}{0,61}             & \multicolumn{1}{c|}{0,79}           & \multicolumn{1}{c|}{0,86}              & \multicolumn{1}{c|}{0,89}              & \multicolumn{1}{c|}{0,49}             & 0,75                                \\ \cline{2-10} 
\multicolumn{1}{|c|}{}                                  & Weighted Avg F1                      & \multicolumn{1}{c|}{}                  & \multicolumn{1}{c|}{}                  & \multicolumn{1}{c|}{}                 & \multicolumn{1}{c|}{0,84}           & \multicolumn{1}{c|}{}                  & \multicolumn{1}{c|}{}                  & \multicolumn{1}{c|}{}                 & 0,81                                \\ \cline{2-10} 
\multicolumn{1}{|c|}{}                                  & Accuracy                             & \multicolumn{1}{c|}{}                  & \multicolumn{1}{c|}{}                  & \multicolumn{1}{c|}{}                 & \multicolumn{1}{c|}{0,85}           & \multicolumn{1}{c|}{}                  & \multicolumn{1}{c|}{}                  & \multicolumn{1}{c|}{}                 & 0,83                                \\ \hline
\multicolumn{1}{|c|}{\multirow{5}{*}{Few Shot}}         & Precision                            & \multicolumn{1}{c|}{0,88}              & \multicolumn{1}{c|}{0,89}              & \multicolumn{1}{c|}{0,66}             & \multicolumn{1}{c|}{0,81}           & \multicolumn{1}{c|}{0,83}              & \multicolumn{1}{c|}{0,87}              & \multicolumn{1}{c|}{0,76}             & 0,82                                \\ \cline{2-10} 
\multicolumn{1}{|c|}{}                                  & Recall                               & \multicolumn{1}{c|}{0,88}              & \multicolumn{1}{c|}{0,91}              & \multicolumn{1}{c|}{0,63}             & \multicolumn{1}{c|}{0,81}           & \multicolumn{1}{c|}{0,94}              & \multicolumn{1}{c|}{0,93}              & \multicolumn{1}{c|}{0,41}             & 0,76                                \\ \cline{2-10} 
\multicolumn{1}{|c|}{}                                  & F1 (macro)                           & \multicolumn{1}{c|}{0,88}              & \multicolumn{1}{c|}{0,9}               & \multicolumn{1}{c|}{0,65}             & \multicolumn{1}{c|}{0,81}           & \multicolumn{1}{c|}{0,88}              & \multicolumn{1}{c|}{0,9}               & \multicolumn{1}{c|}{0,53}             & 0,77                                \\ \cline{2-10} 
\multicolumn{1}{|c|}{}                                  & Weighted Avg F1                      & \multicolumn{1}{c|}{}                  & \multicolumn{1}{c|}{}                  & \multicolumn{1}{c|}{}                 & \multicolumn{1}{c|}{0,85}           & \multicolumn{1}{c|}{}                  & \multicolumn{1}{c|}{}                  & \multicolumn{1}{c|}{}                 & 0,82                                \\ \cline{2-10} 
\multicolumn{1}{|c|}{}                                  & Accuracy                             & \multicolumn{1}{c|}{}                  & \multicolumn{1}{c|}{}                  & \multicolumn{1}{c|}{}                 & \multicolumn{1}{c|}{0,85}           & \multicolumn{1}{c|}{}                  & \multicolumn{1}{c|}{}                  & \multicolumn{1}{c|}{}                 & 0,84                                \\ \hline
\multicolumn{1}{|c|}{\multirow{5}{*}{CoT}}              & Precision                            & \multicolumn{1}{c|}{0,83}              & \multicolumn{1}{c|}{0,9}               & \multicolumn{1}{c|}{0,63}             & \multicolumn{1}{c|}{0,79}           & \multicolumn{1}{c|}{0,87}              & \multicolumn{1}{c|}{0,89}              & \multicolumn{1}{c|}{0,58}             & 0,78                                \\ \cline{2-10} 
\multicolumn{1}{|c|}{}                                  & Recall                               & \multicolumn{1}{c|}{0,92}              & \multicolumn{1}{c|}{0,9}               & \multicolumn{1}{c|}{0,49}             & \multicolumn{1}{c|}{0,77}           & \multicolumn{1}{c|}{0,88}              & \multicolumn{1}{c|}{0,9}               & \multicolumn{1}{c|}{0,56}             & 0,78                                \\ \cline{2-10} 
\multicolumn{1}{|c|}{}                                  & F1 (macro)                           & \multicolumn{1}{c|}{0,87}              & \multicolumn{1}{c|}{0,9}               & \multicolumn{1}{c|}{0,55}             & \multicolumn{1}{c|}{0,77}           & \multicolumn{1}{c|}{0,88}              & \multicolumn{1}{c|}{0,9}               & \multicolumn{1}{c|}{0,57}             & 0,78                                \\ \cline{2-10} 
\multicolumn{1}{|c|}{}                                  & Weighted Avg F1                      & \multicolumn{1}{c|}{}                  & \multicolumn{1}{c|}{}                  & \multicolumn{1}{c|}{}                 & \multicolumn{1}{c|}{0,83}           & \multicolumn{1}{c|}{}                  & \multicolumn{1}{c|}{}                  & \multicolumn{1}{c|}{}                 & 0,83                                \\ \cline{2-10} 
\multicolumn{1}{|c|}{}                                  & Accuracy                             & \multicolumn{1}{c|}{}                  & \multicolumn{1}{c|}{}                  & \multicolumn{1}{c|}{}                 & \multicolumn{1}{c|}{0,83}           & \multicolumn{1}{c|}{}                  & \multicolumn{1}{c|}{}                  & \multicolumn{1}{c|}{}                 & 0,83                                \\ \hline
\multicolumn{1}{|c|}{\multirow{5}{*}{ZeroShot-CoT}}     & Precision                            & \multicolumn{1}{c|}{0,81}              & \multicolumn{1}{c|}{0,85}              & \multicolumn{1}{c|}{0,69}             & \multicolumn{1}{c|}{0,78}           & \multicolumn{1}{c|}{0,83}              & \multicolumn{1}{c|}{0,83}              & \multicolumn{1}{c|}{0,59}             & 0,75                                \\ \cline{2-10} 
\multicolumn{1}{|c|}{}                                  & Recall                               & \multicolumn{1}{c|}{0,93}              & \multicolumn{1}{c|}{0,94}              & \multicolumn{1}{c|}{0,29}             & \multicolumn{1}{c|}{0,72}           & \multicolumn{1}{c|}{0,92}              & \multicolumn{1}{c|}{0,92}              & \multicolumn{1}{c|}{0,29}             & 0,71                                \\ \cline{2-10} 
\multicolumn{1}{|c|}{}                                  & F1 (macro)                           & \multicolumn{1}{c|}{0,87}              & \multicolumn{1}{c|}{0,9}               & \multicolumn{1}{c|}{0,4}              & \multicolumn{1}{c|}{0,72}           & \multicolumn{1}{c|}{0,87}              & \multicolumn{1}{c|}{0,87}              & \multicolumn{1}{c|}{0,39}             & 0,71                                \\ \cline{2-10} 
\multicolumn{1}{|c|}{}                                  & Weighted Avg F1                      & \multicolumn{1}{c|}{}                  & \multicolumn{1}{c|}{}                  & \multicolumn{1}{c|}{}                 & \multicolumn{1}{c|}{0,80}           & \multicolumn{1}{c|}{}                  & \multicolumn{1}{c|}{}                  & \multicolumn{1}{c|}{}                 & 0,79                                \\ \cline{2-10} 
\multicolumn{1}{|c|}{}                                  & Accuracy                             & \multicolumn{1}{c|}{}                  & \multicolumn{1}{c|}{}                  & \multicolumn{1}{c|}{}                 & \multicolumn{1}{c|}{0,82}           & \multicolumn{1}{c|}{}                  & \multicolumn{1}{c|}{}                  & \multicolumn{1}{c|}{}                 & 0,81                                \\ \hline
\multicolumn{1}{|c|}{\multirow{5}{*}{Self-Consistency}} & Precision                            & \multicolumn{1}{c|}{0,83}              & \multicolumn{1}{c|}{0,88}              & \multicolumn{1}{c|}{0,62}             & \multicolumn{1}{c|}{0,78}           & \multicolumn{1}{c|}{0,87}              & \multicolumn{1}{c|}{0,89}              & \multicolumn{1}{c|}{0,59}             & 0,78                                \\ \cline{2-10} 
\multicolumn{1}{|c|}{}                                  & Recall                               & \multicolumn{1}{c|}{0,91}              & \multicolumn{1}{c|}{0,89}              & \multicolumn{1}{c|}{0,48}             & \multicolumn{1}{c|}{0,76}           & \multicolumn{1}{c|}{0,89}              & \multicolumn{1}{c|}{0,9}               & \multicolumn{1}{c|}{0,55}             & 0,78                                \\ \cline{2-10} 
\multicolumn{1}{|c|}{}                                  & F1 (macro)                           & \multicolumn{1}{c|}{0,87}              & \multicolumn{1}{c|}{0,88}              & \multicolumn{1}{c|}{0,54}             & \multicolumn{1}{c|}{0,76}           & \multicolumn{1}{c|}{0,88}              & \multicolumn{1}{c|}{0,89}              & \multicolumn{1}{c|}{0,57}             & 0,78                                \\ \cline{2-10} 
\multicolumn{1}{|c|}{}                                  & Weighted Avg F1                      & \multicolumn{1}{c|}{}                  & \multicolumn{1}{c|}{}                  & \multicolumn{1}{c|}{}                 & \multicolumn{1}{c|}{0,82}           & \multicolumn{1}{c|}{}                  & \multicolumn{1}{c|}{}                  & \multicolumn{1}{c|}{}                 & 0,83                                \\ \cline{2-10} 
\multicolumn{1}{|c|}{}                                  & Accuracy                             & \multicolumn{1}{c|}{}                  & \multicolumn{1}{c|}{}                  & \multicolumn{1}{c|}{}                 & \multicolumn{1}{c|}{0,82}           & \multicolumn{1}{c|}{}                  & \multicolumn{1}{c|}{}                  & \multicolumn{1}{c|}{}                 & 0,83                                \\ \hline
\end{tabular}%
}
\caption{All results ABSA}
\label{tab:Result_absa}
\end{table}

\clearpage
\subsection{Results irony-detection} \label{appendix_section_results_irony_detection}

\begin{table}[!ht]
\centering
\resizebox{0.85\textwidth}{!}{%
\begin{tabular}{cl|cccccc|}
\cline{3-8}
\multicolumn{1}{l}{\textbf{}}                           &                                      & \multicolumn{6}{c|}{\textbf{SemEval-2018}}                                                                                                                                                                                                    \\ \cline{3-8} 
\multicolumn{1}{l}{}                                    &                                      & \multicolumn{3}{c|}{\textbf{GPT}}                                                                                     & \multicolumn{3}{c|}{\textbf{GEMINI}}                                                                                  \\ \hline
\multicolumn{1}{|c|}{\textbf{PE Approach}}              & \multicolumn{1}{c|}{\textbf{Metric}} & \multicolumn{1}{l|}{\textbf{Negative}} & \multicolumn{1}{l|}{\textbf{Positive}} & \multicolumn{1}{l|}{\textbf{Total}} & \multicolumn{1}{l|}{\textbf{Negative}} & \multicolumn{1}{l|}{\textbf{Positive}} & \multicolumn{1}{l|}{\textbf{Total}} \\ \hline
\multicolumn{1}{|c|}{\multirow{5}{*}{BASE}}             & Precision                            & \multicolumn{1}{c|}{0,76}              & \multicolumn{1}{c|}{0,72}              & \multicolumn{1}{c|}{0,74}           & \multicolumn{1}{c|}{0,84}              & \multicolumn{1}{c|}{0,52}              & 0,68                                \\ \cline{2-8} 
\multicolumn{1}{|c|}{}                                  & Recall                               & \multicolumn{1}{c|}{0,67}              & \multicolumn{1}{c|}{0,8}               & \multicolumn{1}{c|}{0,74}           & \multicolumn{1}{c|}{0,06}              & \multicolumn{1}{c|}{0,99}              & 0,53                                \\ \cline{2-8} 
\multicolumn{1}{|c|}{}                                  & F1 (macro)                           & \multicolumn{1}{c|}{0,71}              & \multicolumn{1}{c|}{0,75}              & \multicolumn{1}{c|}{0,73}           & \multicolumn{1}{c|}{0,12}              & \multicolumn{1}{c|}{0,69}              & 0,41                                \\ \cline{2-8} 
\multicolumn{1}{|c|}{}                                  & Weighted Avg F1                      & \multicolumn{1}{c|}{}                  & \multicolumn{1}{c|}{}                  & \multicolumn{1}{c|}{0,73}           & \multicolumn{1}{c|}{}                  & \multicolumn{1}{c|}{}                  & 0,41                                \\ \cline{2-8} 
\multicolumn{1}{|c|}{}                                  & Accuracy                             & \multicolumn{1}{c|}{}                  & \multicolumn{1}{c|}{}                  & \multicolumn{1}{c|}{0,74}           & \multicolumn{1}{c|}{}                  & \multicolumn{1}{c|}{}                  & 0,54                                \\ \hline
\multicolumn{1}{|c|}{\multirow{5}{*}{One Shot}}         & Precision                            & \multicolumn{1}{c|}{0,78}              & \multicolumn{1}{c|}{0,7}               & \multicolumn{1}{c|}{0,74}           & \multicolumn{1}{c|}{0,81}              & \multicolumn{1}{c|}{0,58}              & 0,70                                \\ \cline{2-8} 
\multicolumn{1}{|c|}{}                                  & Recall                               & \multicolumn{1}{c|}{0,62}              & \multicolumn{1}{c|}{0,83}              & \multicolumn{1}{c|}{0,73}           & \multicolumn{1}{c|}{0,28}              & \multicolumn{1}{c|}{0,94}              & 0,61                                \\ \cline{2-8} 
\multicolumn{1}{|c|}{}                                  & F1 (macro)                           & \multicolumn{1}{c|}{0,69}              & \multicolumn{1}{c|}{0,76}              & \multicolumn{1}{c|}{0,73}           & \multicolumn{1}{c|}{0,42}              & \multicolumn{1}{c|}{0,71}              & 0,57                                \\ \cline{2-8} 
\multicolumn{1}{|c|}{}                                  & Weighted Avg F1                      & \multicolumn{1}{c|}{}                  & \multicolumn{1}{c|}{}                  & \multicolumn{1}{c|}{0,72}           & \multicolumn{1}{c|}{}                  & \multicolumn{1}{c|}{}                  & 0,57                                \\ \cline{2-8} 
\multicolumn{1}{|c|}{}                                  & Accuracy                             & \multicolumn{1}{c|}{}                  & \multicolumn{1}{c|}{}                  & \multicolumn{1}{c|}{0,73}           & \multicolumn{1}{c|}{}                  & \multicolumn{1}{c|}{}                  & 0,62                                \\ \hline
\multicolumn{1}{|c|}{\multirow{5}{*}{Few Shot}}         & Precision                            & \multicolumn{1}{c|}{0,76}              & \multicolumn{1}{c|}{0,75}              & \multicolumn{1}{c|}{0,76}           & \multicolumn{1}{c|}{0,78}              & \multicolumn{1}{c|}{0,57}              & 0,68                                \\ \cline{2-8} 
\multicolumn{1}{|c|}{}                                  & Recall                               & \multicolumn{1}{c|}{0,73}              & \multicolumn{1}{c|}{0,78}              & \multicolumn{1}{c|}{0,76}           & \multicolumn{1}{c|}{0,28}              & \multicolumn{1}{c|}{0,93}              & 0,61                                \\ \cline{2-8} 
\multicolumn{1}{|c|}{}                                  & F1 (macro)                           & \multicolumn{1}{c|}{0,74}              & \multicolumn{1}{c|}{0,77}              & \multicolumn{1}{c|}{0,76}           & \multicolumn{1}{c|}{0,41}              & \multicolumn{1}{c|}{0,71}              & 0,56                                \\ \cline{2-8} 
\multicolumn{1}{|c|}{}                                  & Weighted Avg F1                      & \multicolumn{1}{c|}{}                  & \multicolumn{1}{c|}{}                  & \multicolumn{1}{c|}{0,76}           & \multicolumn{1}{c|}{}                  & \multicolumn{1}{c|}{}                  & 0,56                                \\ \cline{2-8} 
\multicolumn{1}{|c|}{}                                  & Accuracy                             & \multicolumn{1}{c|}{}                  & \multicolumn{1}{c|}{}                  & \multicolumn{1}{c|}{0,76}           & \multicolumn{1}{c|}{}                  & \multicolumn{1}{c|}{}                  & 0,61                                \\ \hline
\multicolumn{1}{|c|}{\multirow{5}{*}{CoT}}              & Precision                            & \multicolumn{1}{c|}{0,77}              & \multicolumn{1}{c|}{0,63}              & \multicolumn{1}{c|}{0,70}           & \multicolumn{1}{c|}{0,75}              & \multicolumn{1}{c|}{0,59}              & 0,67                                \\ \cline{2-8} 
\multicolumn{1}{|c|}{}                                  & Recall                               & \multicolumn{1}{c|}{0,47}              & \multicolumn{1}{c|}{0,87}              & \multicolumn{1}{c|}{0,67}           & \multicolumn{1}{c|}{0,37}              & \multicolumn{1}{c|}{0,88}              & 0,63                                \\ \cline{2-8} 
\multicolumn{1}{|c|}{}                                  & F1 (macro)                           & \multicolumn{1}{c|}{0,58}              & \multicolumn{1}{c|}{0,73}              & \multicolumn{1}{c|}{0,66}           & \multicolumn{1}{c|}{0,49}              & \multicolumn{1}{c|}{0,71}              & 0,60                                \\ \cline{2-8} 
\multicolumn{1}{|c|}{}                                  & Weighted Avg F1                      & \multicolumn{1}{c|}{}                  & \multicolumn{1}{c|}{}                  & \multicolumn{1}{c|}{0,66}           & \multicolumn{1}{c|}{}                  & \multicolumn{1}{c|}{}                  & 0,60                                \\ \cline{2-8} 
\multicolumn{1}{|c|}{}                                  & Accuracy                             & \multicolumn{1}{c|}{}                  & \multicolumn{1}{c|}{}                  & \multicolumn{1}{c|}{0,67}           & \multicolumn{1}{c|}{}                  & \multicolumn{1}{c|}{}                  & 0,63                                \\ \hline
\multicolumn{1}{|c|}{\multirow{5}{*}{ZeroShot-CoT}}     & Precision                            & \multicolumn{1}{c|}{0,77}              & \multicolumn{1}{c|}{0,6}               & \multicolumn{1}{c|}{0,69}           & \multicolumn{1}{c|}{0,79}              & \multicolumn{1}{c|}{0,53}              & 0,66                                \\ \cline{2-8} 
\multicolumn{1}{|c|}{}                                  & Recall                               & \multicolumn{1}{c|}{0,38}              & \multicolumn{1}{c|}{0,89}              & \multicolumn{1}{c|}{0,64}           & \multicolumn{1}{c|}{0,11}              & \multicolumn{1}{c|}{0,97}              & 0,54                                \\ \cline{2-8} 
\multicolumn{1}{|c|}{}                                  & F1 (macro)                           & \multicolumn{1}{c|}{0,51}              & \multicolumn{1}{c|}{0,72}              & \multicolumn{1}{c|}{0,62}           & \multicolumn{1}{c|}{0,19}              & \multicolumn{1}{c|}{0,69}              & 0,44                                \\ \cline{2-8} 
\multicolumn{1}{|c|}{}                                  & Weighted Avg F1                      & \multicolumn{1}{c|}{}                  & \multicolumn{1}{c|}{}                  & \multicolumn{1}{c|}{0,62}           & \multicolumn{1}{c|}{}                  & \multicolumn{1}{c|}{}                  & 0,44                                \\ \cline{2-8} 
\multicolumn{1}{|c|}{}                                  & Accuracy                             & \multicolumn{1}{c|}{}                  & \multicolumn{1}{c|}{}                  & \multicolumn{1}{c|}{0,64}           & \multicolumn{1}{c|}{}                  & \multicolumn{1}{c|}{}                  & 0,55                                \\ \hline
\multicolumn{1}{|c|}{\multirow{5}{*}{Self-Consistency}} & Precision                            & \multicolumn{1}{c|}{0,72}              & \multicolumn{1}{c|}{0,56}              & \multicolumn{1}{c|}{0,64}           & \multicolumn{1}{c|}{0,77}              & \multicolumn{1}{c|}{0,54}              & 0,66                                \\ \cline{2-8} 
\multicolumn{1}{|c|}{}                                  & Recall                               & \multicolumn{1}{c|}{0,31}              & \multicolumn{1}{c|}{0,88}              & \multicolumn{1}{c|}{0,60}           & \multicolumn{1}{c|}{0,13}              & \multicolumn{1}{c|}{0,96}              & 0,55                                \\ \cline{2-8} 
\multicolumn{1}{|c|}{}                                  & F1 (macro)                           & \multicolumn{1}{c|}{0,43}              & \multicolumn{1}{c|}{0,68}              & \multicolumn{1}{c|}{0,56}           & \multicolumn{1}{c|}{0,22}              & \multicolumn{1}{c|}{0,69}              & 0,46                                \\ \cline{2-8} 
\multicolumn{1}{|c|}{}                                  & Weighted Avg F1                      & \multicolumn{1}{c|}{}                  & \multicolumn{1}{c|}{}                  & \multicolumn{1}{c|}{0,56}           & \multicolumn{1}{c|}{}                  & \multicolumn{1}{c|}{}                  & 0,44                                \\ \cline{2-8} 
\multicolumn{1}{|c|}{}                                  & Accuracy                             & \multicolumn{1}{c|}{}                  & \multicolumn{1}{c|}{}                  & \multicolumn{1}{c|}{0,59}           & \multicolumn{1}{c|}{}                  & \multicolumn{1}{c|}{}                  & 0,55                                \\ \hline
\end{tabular}%
}
\caption{All results irony-detection}
\label{tab:Result_irony}
\end{table}

\setcounter{table}{0}
\renewcommand\thetable{D.\arabic{table}}

\setcounter{figure}{0}
\renewcommand\thefigure{D.\arabic{figure}}

\clearpage
\section{Confusion matrices}\label{appendix:conf_m}

\subsection{One-shot prompting} \label{appendix_section_one_shot}

\begin{figure}[htbp]
    \centering
    \begin{subfigure}[t]{0.45\textwidth}
        \centering
        \includegraphics[width=\textwidth]{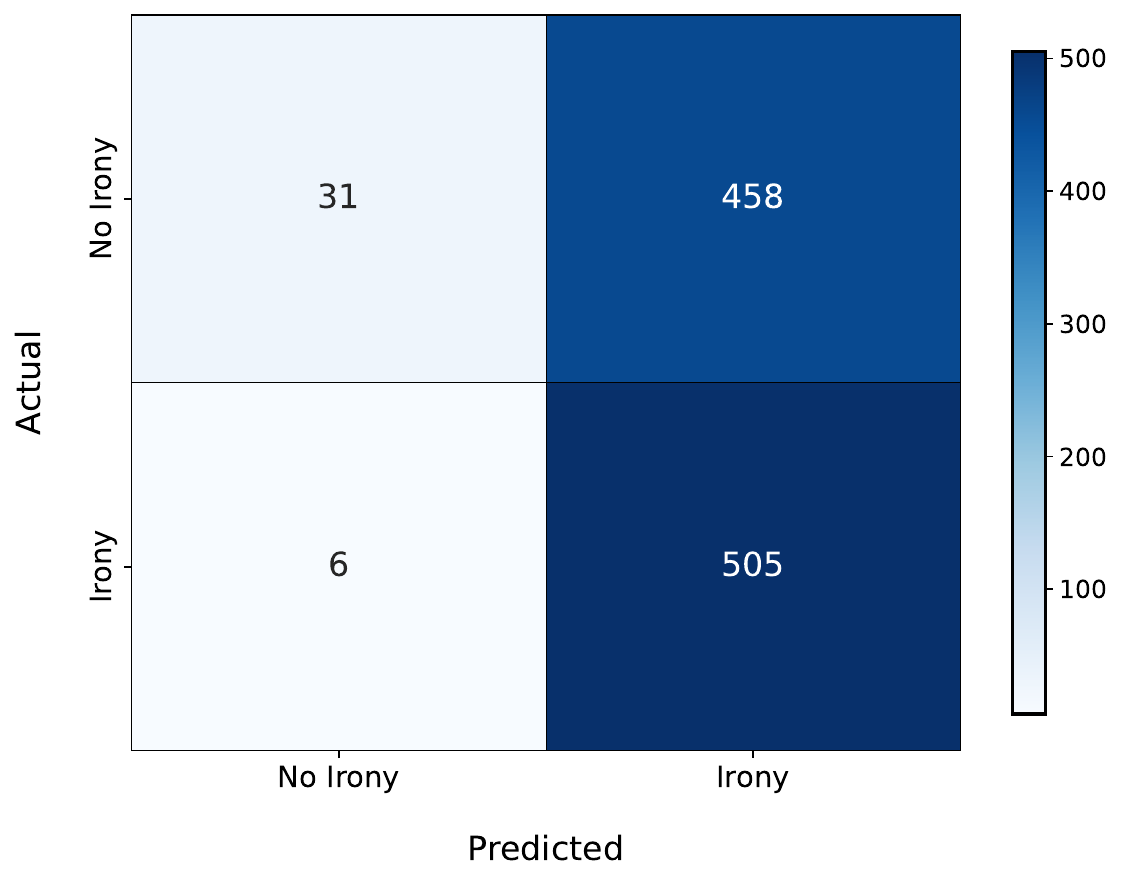}
        \caption{Baseline prompting}
        \label{fig:cm_baseline}
    \end{subfigure}
    \hfill
    \begin{subfigure}[t]{0.45\textwidth}
        \centering
        \includegraphics[width=\textwidth]{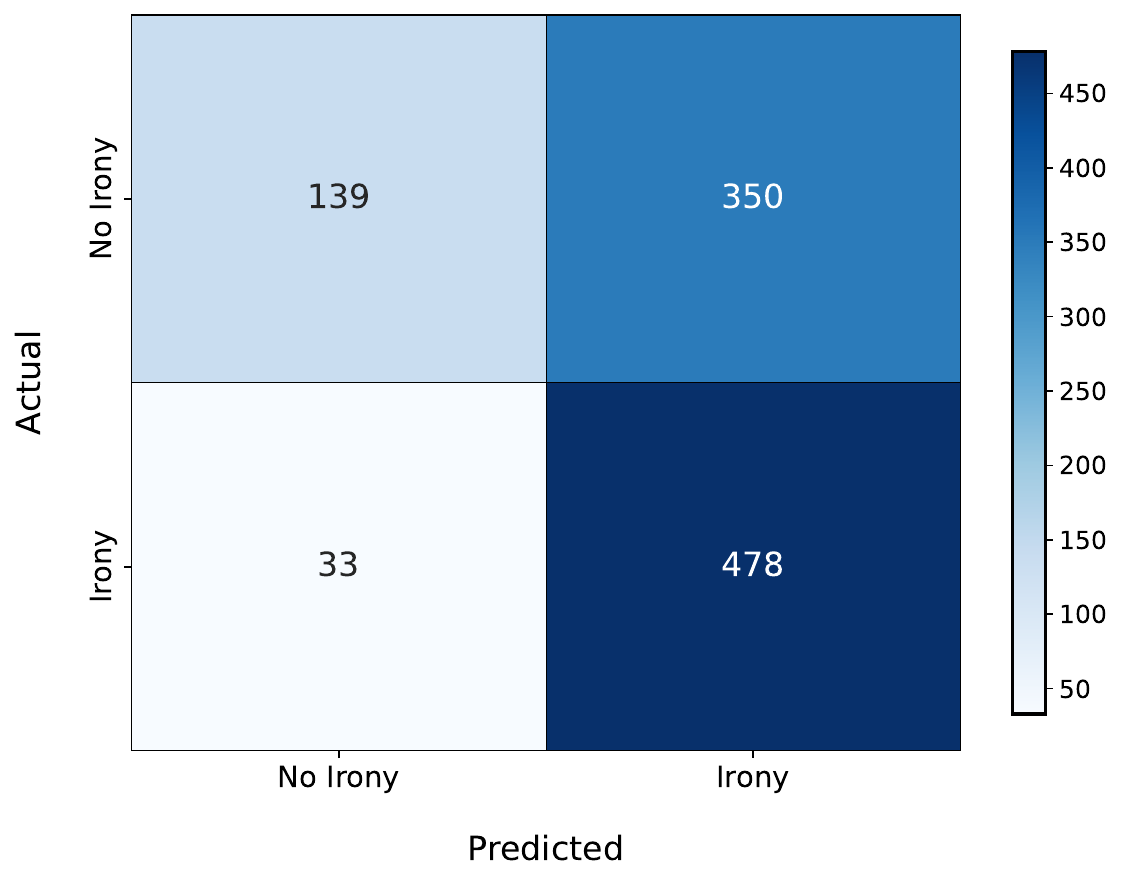}
        \caption{One-shot prompting}
        \label{fig:cm_one_shot}
    \end{subfigure}
    \caption{Confusion matrix baseline vs. one-shot irony detection (gemini-flash1.5).}
    \label{fig:cm_irony1}
\end{figure}

In Figure \ref{fig:cm_irony1}, the confusion matrix for the irony detection task under the gemini-1.5-flash model using one-shot prompting is presented. The results indicate that even when the positive class (i.e., “irony”) is used as the single example, the model’s performance in detecting the negative class (i.e., “no irony”) improves.\\


\begin{figure}[htbp]
    \centering
    \begin{subfigure}[t]{0.45\textwidth}
        \centering
        \includegraphics[width=\textwidth]{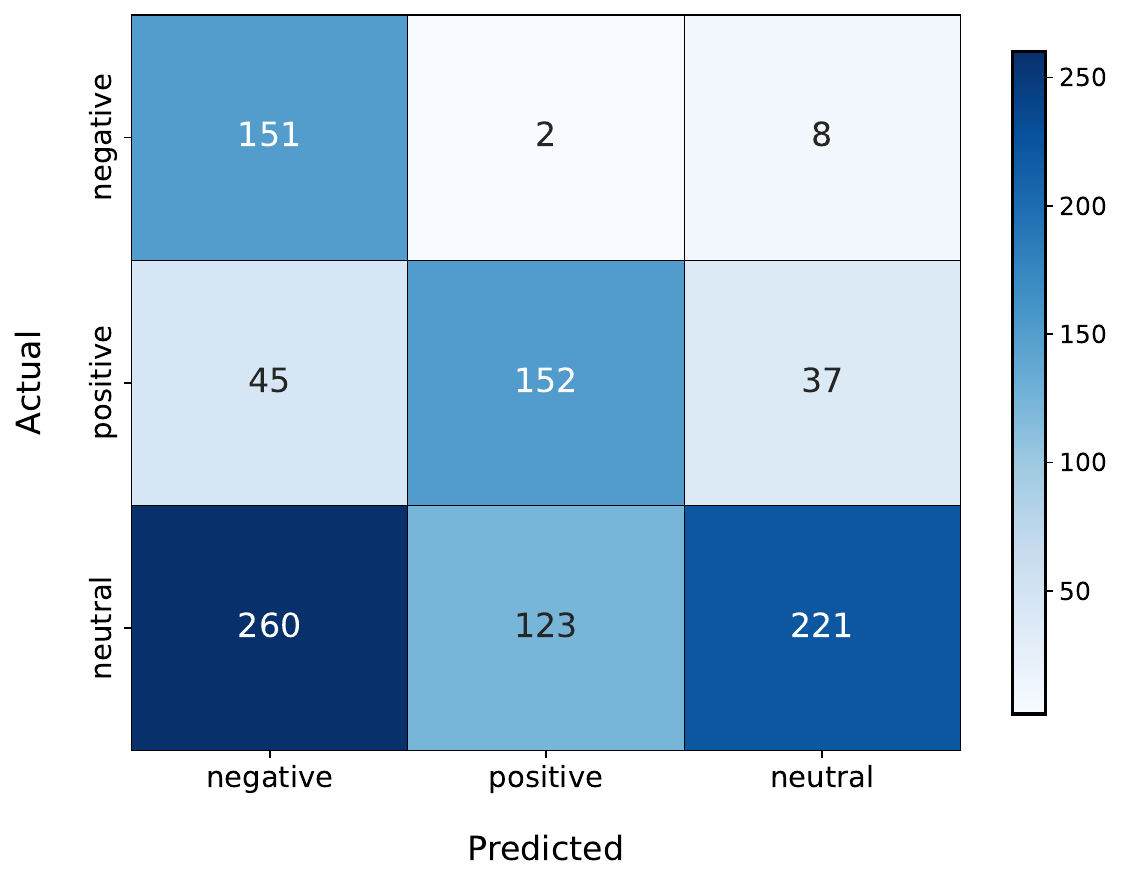}
        \caption{Baseline prompting}
        \label{fig:cm_baseline_sb10k}
    \end{subfigure}
    \hfill
    \begin{subfigure}[t]{0.45\textwidth}
        \centering
        \includegraphics[width=\textwidth]{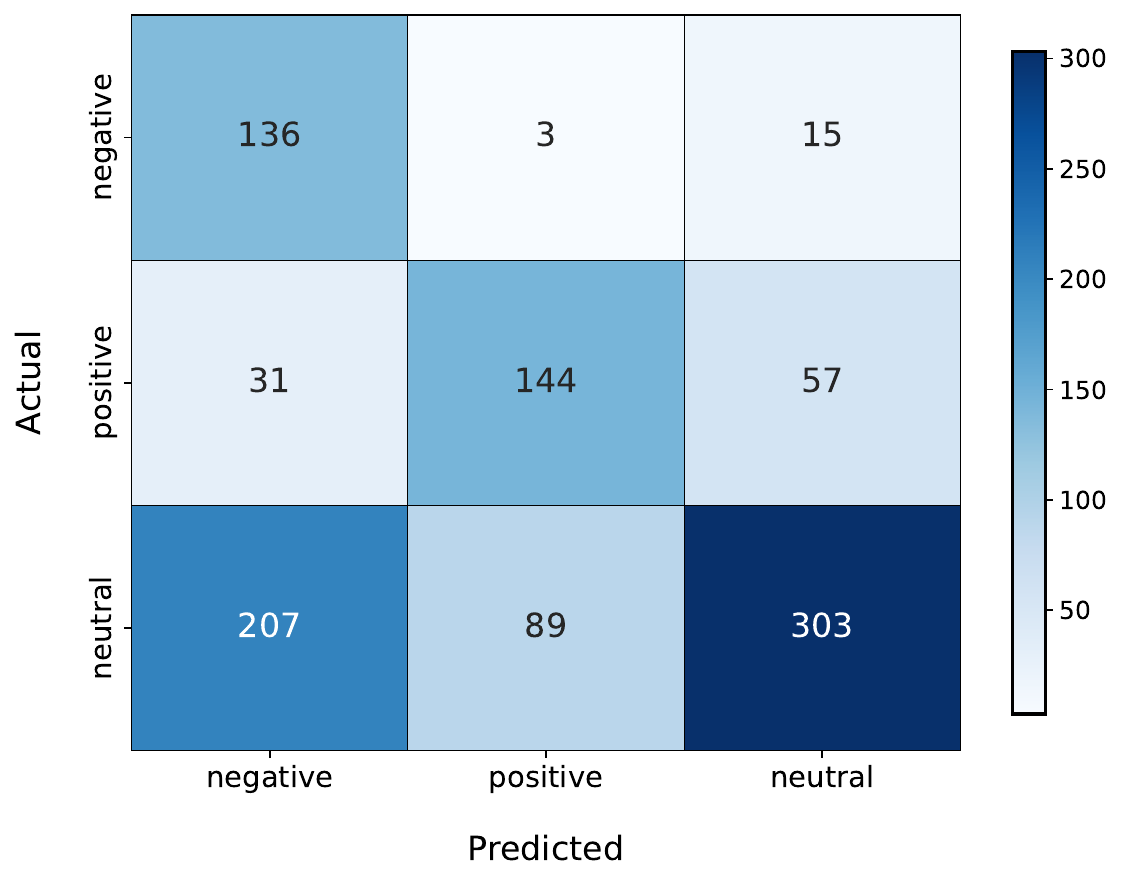}
        \caption{One-shot prompting}
        \label{fig:cm_one_shot_sb10k}
    \end{subfigure}
    \caption{Confusion matrix baseline vs. one-shot SB10k (gemini-flash-1.5).}
    \label{fig:cm_sb10k}
\end{figure}

In Figure \ref{fig:cm_sb10k}, it can be observed that the performance for the neutral class improves significantly when a single example of that class is provided using one-shot prompting.

\clearpage
\subsection{CoT prompting} \label{appendix:CoT}

\begin{figure}[htbp]
    \centering
    \begin{subfigure}[t]{0.45\textwidth}
        \centering
        \includegraphics[width=\textwidth]{confusion_matrix_base_irony_gemini.pdf}
        \caption{Baseline prompting}
        \label{fig:cm_baseline_gemini}
    \end{subfigure}
    \hfill
    \begin{subfigure}[t]{0.45\textwidth}
        \centering
        \includegraphics[width=\textwidth]{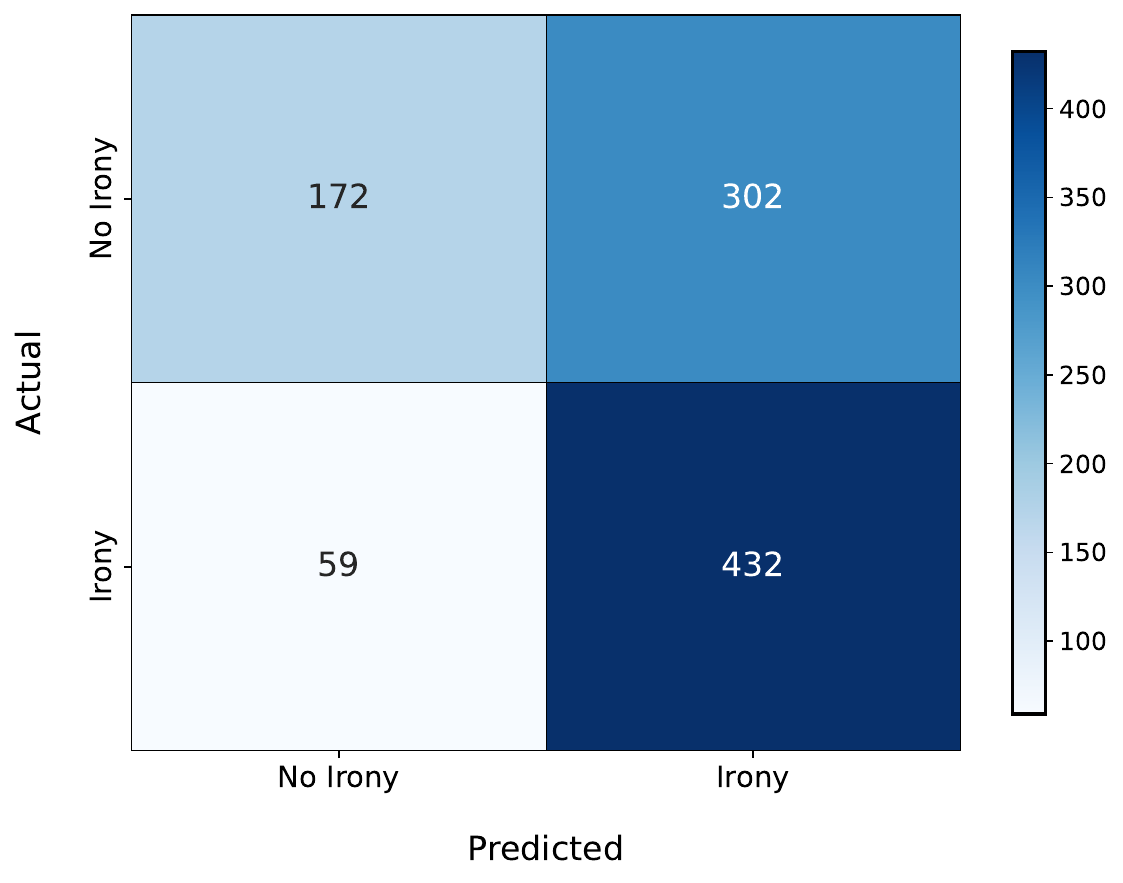}
        \caption{CoT prompting}
        \label{fig:cm_CoT_gemini}
    \end{subfigure}
    \caption{Confusion matrix baseline vs. CoT irony detection (gemini-flash-1.5)}
    \label{fig:cm_irony_CoT_gemini}
\end{figure}

Figure \ref{fig:cm_irony_CoT_gemini} shows that the recall for the negative class (“no irony”) is extremely low (\textit{0.06}) in the zero-shot baseline, as most texts are falsely classified as ironic. In contrast, the CoT approach improves the recall for this class to \textit{0.38}, indicating better distinction between ironic and non-ironic content.

\subsection{Self-consistency prompting} \label{appendix:SelfCons}

\begin{figure}[htbp]
    \centering
    \begin{subfigure}[t]{0.45\textwidth}
        \centering
        \includegraphics[width=\textwidth]{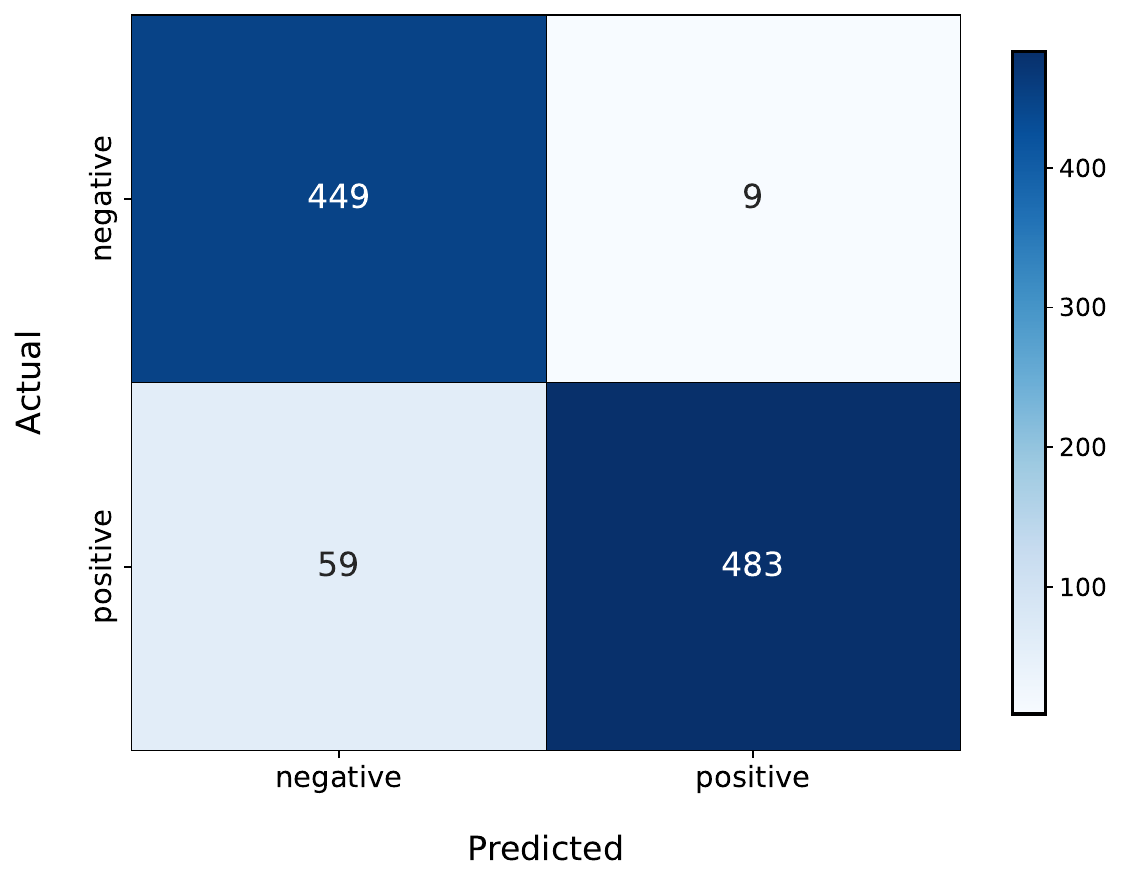}
        \caption{Few-shot prompting}
        \label{fig:cm_FS_SST2}
    \end{subfigure}
    \hfill
    \begin{subfigure}[t]{0.45\textwidth}
        \centering
        \includegraphics[width=\textwidth]{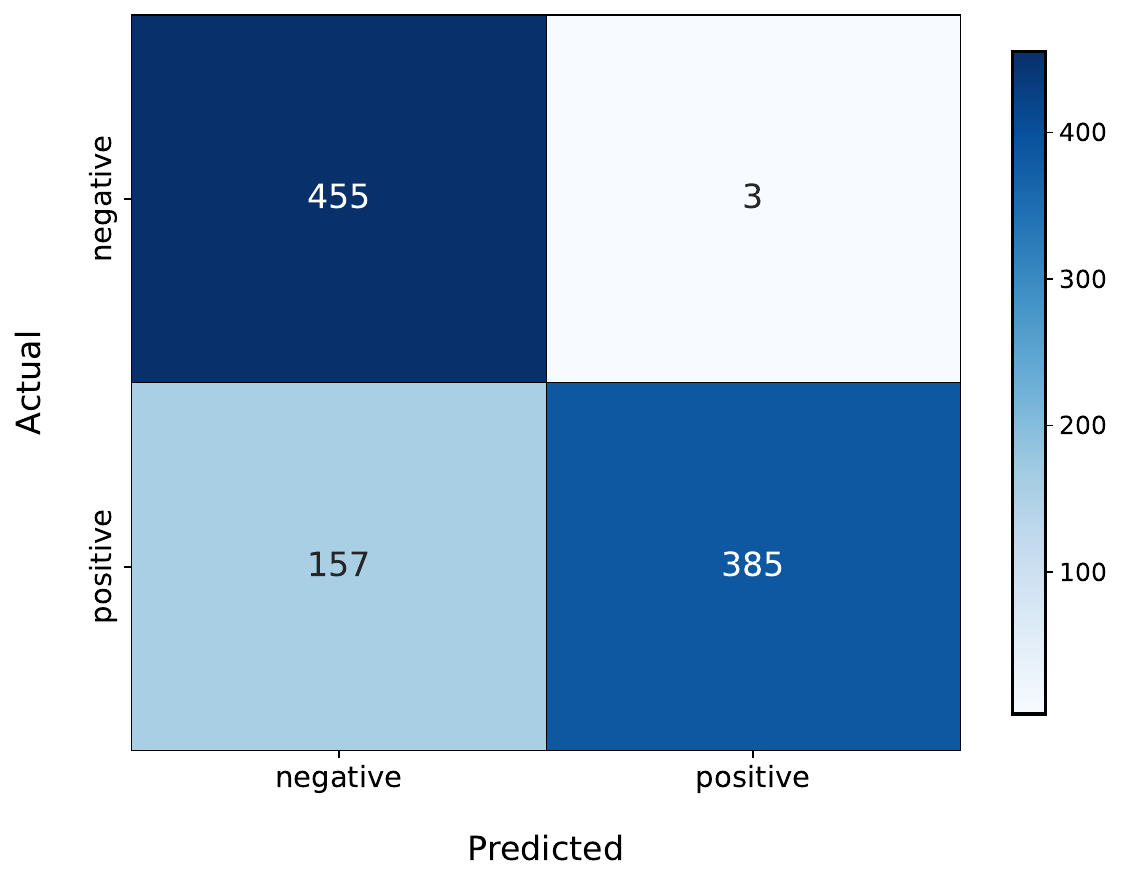}
        \caption{Self-consistency prompting}
        \label{fig:cm_SC}
    \end{subfigure}
    \caption{Confusion matrix few-shot vs. self-consistency SST2 (GPT-4o-mini).}
    \label{fig:cm_SST2_GPT}
\end{figure}

Figure \ref{fig:cm_SST2_GPT} compares the confusion matrices of the best-performing (few-shot) and worst-performing (self-consistency) GPT-4o-mini approaches. The self-consistency method shows a high number of false negatives, leading to low precision for the negative class. Interestingly, the model consistently misclassifies with apparent confidence, despite majority voting.

\clearpage
\twocolumn




\printbibliography

\end{document}